\documentclass{article}
\usepackage[utf8]{inputenc}
\usepackage[T1]{fontenc}
\usepackage{main}
\usepackage{microtype}
\usepackage{graphicx}
\usepackage{times}
\usepackage{amsmath}
\usepackage{amssymb}
\usepackage{amsfonts}
\usepackage{amsthm}
\usepackage{bbm}
\usepackage{bm}
\usepackage{mathtools}
\usepackage{nicefrac}
\usepackage{enumitem}
\usepackage{booktabs}
\usepackage[table]{xcolor}
\usepackage{url}
\definecolor{mydarkblue}{rgb}{0,0.08,0.45}
\usepackage[colorlinks=true,linkcolor=mydarkblue,citecolor=mydarkblue,filecolor=mydarkblue,urlcolor=mydarkblue]{hyperref}
\usepackage{fancyhdr}

\usepackage{comment}
\usepackage{dsfont}
\usepackage{float}
\usepackage{multirow}
\usepackage{makecell}
\usepackage{hhline}
\usepackage{colortbl}
\usepackage{algorithm}
\usepackage{algpseudocode}
\usepackage[most]{tcolorbox}
\usepackage{etoc}
\usepackage{placeins}

\definecolor{gred}{RGB}{250, 210, 207}
\definecolor{coolblue1}{rgb}{0.91, 0.94, 0.98}
\definecolor{coolblue2}{rgb}{0.76, 0.85, 0.94}
\definecolor{coolblue3}{rgb}{0.54, 0.72, 0.87}
\definecolor{coolblue4}{rgb}{1, 1, 1}

\usepackage{xspace}
\usepackage{cleveref}
\usepackage[]{multicol, multirow}
\usepackage[most]{tcolorbox}
\usepackage{etoc}
\usepackage[table]{xcolor}

\tcbuselibrary{listingsutf8}
\tcbuselibrary{listingsutf8,breakable}

\newtcolorbox[auto counter]{observation}[1][]{
  colback=black!5!white,
  colframe=black!70!white,
  fonttitle=\bfseries,
  title=Observation~\thetcbcounter,
  enhanced,
  boxrule=0.6pt,
  left=1mm,right=1mm,top=1mm,bottom=1mm,
  #1
}

\newtcolorbox[auto counter]{takeaway}[1][]{
  colback=teal!3!white,
  colframe=teal!55!black,
  fonttitle=\bfseries,
  title=Takeaway~\thetcbcounter,
  enhanced,
  boxrule=0.5pt,
  left=1mm,right=1mm,top=1mm,bottom=1mm,
  #1
}

\newtcolorbox[auto counter]{practicalguidance}[1][]{
  colback=cyan!3!white,
  colframe=cyan!60!black,
  fonttitle=\bfseries,
  title=Practical Guidance~\thetcbcounter,
  enhanced,
  boxrule=0.5pt,
  left=1mm,right=1mm,top=1mm,bottom=1mm,
  #1
}

\newtcolorbox[auto counter]{discussion}[1][]{
  colback=violet!4!white,
  colframe=violet!60!black,
  fonttitle=\bfseries,
  title=Discussion~\thetcbcounter,
  enhanced,
  boxrule=0.5pt,
  left=1mm,right=1mm,top=1mm,bottom=1mm,
  #1
}

\allowdisplaybreaks[4]

\definecolor{algogray}{RGB}{130,130,130}
\definecolor{mypurple}{RGB}{120,40,180}
\definecolor{thmblue}{RGB}{40,80,180}
\definecolor{deforange}{RGB}{220,130,40}
\usepackage{etoolbox}

\tcbset{themedboxbase/.style={
  enhanced, breakable,
  boxrule=0.35pt, arc=1mm,
  coltitle=black, fonttitle=\small\sffamily\bfseries,
  attach boxed title to top left={xshift=4mm,yshift*=-1.2mm},
  boxsep=1.5mm, top=1.5mm, bottom=1.5mm, left=1mm, right=1mm,
  before skip=10pt, after skip=10pt
}}

\newtcolorbox{algorithmbox}[1]{%
  themedboxbase,
  colframe=algogray,
  colbacktitle=algogray!20!white,
  colback=algogray!5!white,
  boxed title style={sharp corners, boxrule=0pt,
    top=1pt, bottom=0.5pt, left=4mm, right=4mm,
    borderline={0.5pt}{0pt}{algogray!40!white}},
  title={\textbf{#1}}
}

\tcbset{thmboxstyle/.style={
  themedboxbase,
  colframe=mypurple,
  colbacktitle=mypurple!15!white,
  colback=mypurple!5!white,
  boxed title style={sharp corners, boxrule=0pt,
    top=1pt, bottom=0.5pt, left=4mm, right=4mm,
    borderline={0.5pt}{0pt}{mypurple!40!white}}
}}

\tcbset{theoremboxstyle/.style={
  themedboxbase,
  colframe=thmblue,
  colbacktitle=thmblue!15!white,
  colback=thmblue!5!white,
  boxed title style={sharp corners, boxrule=0pt,
    top=1pt, bottom=0.5pt, left=4mm, right=4mm,
    borderline={0.5pt}{0pt}{thmblue!40!white}}
}}

\tcbset{defboxstyle/.style={
  themedboxbase,
  colframe=deforange,
  colbacktitle=deforange!15!white,
  colback=deforange!5!white,
  boxed title style={sharp corners, boxrule=0pt,
    top=1pt, bottom=0.5pt, left=4mm, right=4mm,
    borderline={0.5pt}{0pt}{deforange!40!white}}
}}

\newtcolorbox{responsebox}[2][]{
    breakable, enhanced,
    colback=cyan!5, colframe=cyan!50!blue,
    coltext=black, coltitle=white,
    fonttitle=\bfseries\rmfamily,
    arc=1mm, boxrule=1pt,
    width=0.95\linewidth, center,
    title=#2, #1
}

\theoremstyle{plain}
\newtheorem{lemma}{Lemma}

\newtheorem{definition}{Definition}

\theoremstyle{remark}

\definecolor{skyblue}{HTML}{4A90D9}

\newcommand{\modelname}{\textsc{LADS}}
\newcommand{\modelnamesimple}{\textsc{LADS-simple}}
\newcommand{\modelnamemath}{\mathrm{LADS}}

\definecolor{jcg}{RGB}{100,160,0}

\definecolor{zz}{RGB}{145, 30, 180}

\title{Lossless Anti-Distillation Sampling}

\author{
    \textbf{Zibo Diao}$^{\dagger}$ \quad
    \textbf{Jingchu Gai}$^{\ddagger}$ \quad
    \textbf{Xinyue Ai}$^{*}$ \quad
    \textbf{Zhang Zhang}$^{*}$ \quad
    \textbf{Zhenyu He}$^{*}$ \quad
    \textbf{Di He}$^{*}$ \\[2pt]
    $^{*}$Peking University \quad $^{\dagger}$Tsinghua University \quad $^{\ddagger}$Carnegie Mellon University \\[2pt]
    Corresponding authors: \texttt{dihe@pku.edu.cn}, \texttt{jgai@andrew.cmu.edu}
}

\begin{document}
\raggedbottom
\sloppy

\maketitle
\thispagestyle{fancy}
\fancyhead{}
\lhead{%
  \raisebox{-0.85cm}[0pt][0pt]{%
    \includegraphics[height=1.0cm]{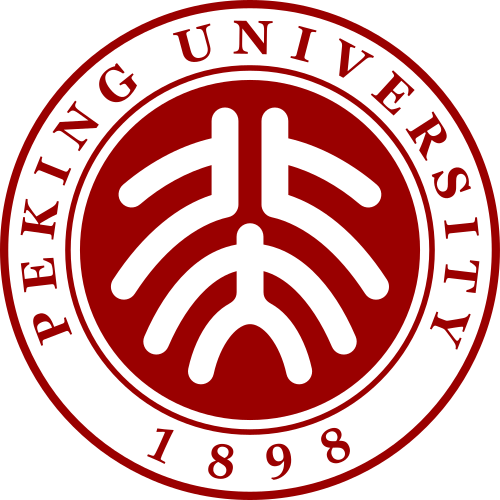}\hspace{4pt}%
    \raisebox{0.28cm}{\sffamily\bfseries\small Peking University}%
  }%
}
\chead{%
  \hspace*{-1.0cm}\raisebox{-0.85cm}[0pt][0pt]{%
    \includegraphics[height=1.0cm]{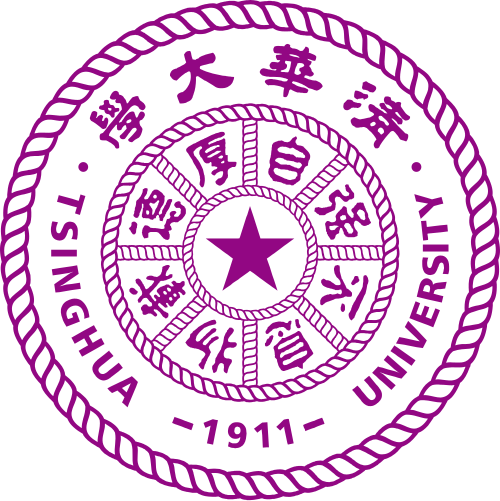}\hspace{4pt}%
    \raisebox{0.28cm}{\sffamily\bfseries\small Tsinghua University}%
  }%
}
\rhead{%
  \raisebox{-0.85cm}[0pt][0pt]{%
    \includegraphics[height=1.0cm,trim=100 20 100 20,clip]{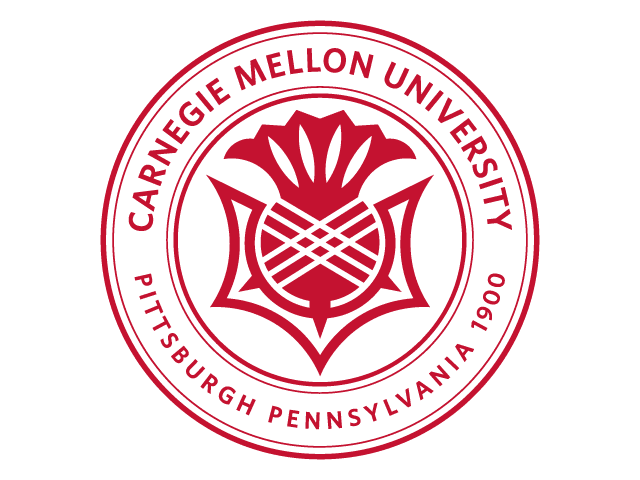}\hspace{4pt}%
    \raisebox{0.28cm}{\sffamily\bfseries\small Carnegie Mellon University}%
  }%
}
\fancyfoot{}
\cfoot{\thepage}
\renewcommand{\headrulewidth}{0pt}
\renewcommand{\footrulewidth}{0pt}
\setlength{\headheight}{36pt}
\setlength{\headsep}{8mm}
\addtolength{\topmargin}{-1.6cm}

\pagestyle{plain}

\begin{abstract}
Frontier commercial generative models face a growing threat from distillation, whereby a distiller harvests generated responses and trains a competing model at drastically lower cost. Existing defenses either modify the generation to degrade distillation performance, sacrificing response quality, or rely on behavioral detection mechanisms that can be readily bypassed through multi-account querying. In this work, we propose Lossless Anti-Distillation Sampling (\modelname{}), which leaves the generation itself unchanged while substantially reducing the effectiveness of distillation. Concretely, \modelname{} controls the latent randomness underlying inference through a coupling mechanism that preserves within-account generation independence while inducing cross-account dependence. By construction, each benign user, who typically holds only a single account, receives the same experience under \modelname{} as they would without any defense, thereby enjoying a lossless experience. However, for a multi-account task-specific distiller, semantically similar queries submitted across different accounts are assigned coupled randomness, inducing dependence in the harvested data and thereby degrading the generalization performance of the distilled model. Using uniform convergence theory, we show that \modelname{} provably degrades the distiller’s generalization gap relative to standard i.i.d.\ sampling. Experiments on image generation, mathematical reasoning, and code generation confirm that \modelname{} substantially degrades the performance of distilled students while preserving exact statistical fidelity for individual users.
\end{abstract}

\setcounter{tocdepth}{2}
\tableofcontents
\FloatBarrier

\section{Introduction}
Commercial AI evolves rapidly. Once a leading company releases a new frontier model, its competitors come under strong pressure to catch up. A well-known  low-risk approach is distillation, in which adversaries harvest large volumes of generated responses by querying the API and then use them to train their own models via supervised fine-tuning or reinforcement fine-tuning. Such a strategy is undesirable for service providers, as it undermines the technical advantage of original innovators and weakens incentives for continued investment in fundamental research.

Given the intense competition among commercial LLM providers, an adversary may need to collect sufficient data and complete fine-tuning within a short period to achieve competitive performance. If the data are collected through a single account, this requires a high query rate, making account-level behavioral detection and restriction relatively straightforward. In practice, a more challenging threat is multi-account distillation, where the adversary controls hundreds of accounts, each designed to behave like a benign user. Existing defenses either degrade benign-user response quality through output-side perturbations~\citep{orekondy2019prediction,ma2021undistillable,savani2025antidistillation}, attribute distillation only post hoc through watermarking~\citep{zhao2023protecting,sander2024watermarking,xu2026antidistillation}, or rely on accurate detection models that are readily circumvented under the multi-account distillation setting~\citep{juuti2019prada,chen2025queen,mei2025defense}. This highlights the need for a defense that is lossless for benign users while remaining effective against multi-account adversaries.

In this paper, we introduce Lossless Anti-Distillation Sampling (\modelname{}). At a high level, \modelname{} views the accesses of different accounts as separate sampling processes and couples the randomness across these processes. The coupling mechanism is carefully designed so that each individual process behaves exactly as it would without the defense, thereby preserving a lossless experience for every individual account, while the samples collected across multiple accounts become statistically dependent, thereby hindering distillation. Concretely, for each generation request \(q\), \modelname{} derives the random seed from the semantic content of the query and the user’s visit count within the corresponding semantic bucket, and generation is then performed using the resulting seed. We show that, for any individual account, the response distribution at each access is identical to that of the original model, while responses across accesses remain independent. Consequently, benign users experience no distortion. At the same time, semantically similar queries submitted across different accounts are coupled through their shared semantics and visit counts. As a result, the samples collected by a multi-account distiller become dependent rather than independent, which reduces sample diversity and consequently degrades the generalization performance of the student model. We support this claim through both a uniform convergence analysis and empirical experiments, as summarized below.


\noindent\textbf{Theoretical Guarantee of \modelname{}.}
We formalize the multi-account threat as a distiller controlling $K$ accounts, each issuing $T$ queries, and minimizing the standard empirical risk defined over the collected data. We consider two main settings: a warm-up unconditional generation setting (\modelnamesimple{}), where the query $q$ is omitted, and a conditional generation setting (\modelname{}), which is more representative of real-world distillation scenarios. Through a uniform convergence analysis, we show that the generalization gap decreases as $\mathcal{O}(1/\sqrt{KT})$ under standard i.i.d.\ sampling, but degrades to $\mathcal{O}(1/\sqrt{T})$ under \modelnamesimple{} (Proposition~\ref{prop:collapse}); under \modelname{}, the generalization gap behaves similarly, depending on the size of the task-associated query space (Theorem~\ref{thm:conditional_main}). In other words, by introducing cross-account coupled noise, additional dummy accounts no longer provide sufficiently informative training signals. As a result, the student model's test error cannot be reduced at the same rate by simply scaling up the number of accounts.

\noindent\textbf{Empirical Validation Across Language and Image Scenarios.}
For image generation, we deploy \modelname{} on diffusion-based EDM2 models. Specifically, we use EDM2-Large (777M) and EDM2-XS (125M) as teachers and EDM2-XXS (31M) as the student. To model the multi-account setting, we distribute ImageNet class queries across $50$ accounts. Under this setup, we introduce a noise-mixing mechanism that interpolates between standard i.i.d.\ sampling and the full \modelname{} scheme, while preserving the exact per-request noise distribution $\mathcal{N}(0,I)$. For language generation, we use Qwen3.5-397B-A17B-FP8 as the teacher, and Llama-3.1-8B, DeepSeek-Math-7B-Base, and Qwen2.5-7B as students. We evaluate distillation quality on MATH, GSM8K and HumanEval. Across both scenarios, we observe that for the student models, FID, validation loss, and task accuracy degrade under \modelname{}, while samples observed by benign users remain of high quality --- confirming that \modelname{} achieves the multi-account distillation defense predicted by our theory under a strict losslessness guarantee.
\section{Related Work}\label{related_work}

\noindent\textbf{Distillation in LLMs.}
Existing work broadly follows two lines. task-specific distillation transfers a targeted capability, such as mathematical reasoning \citep{tong2024dart,pei2025mathfusion}, code generation \citep{luo2023wizardcoder,wei2023magicoder}, or step-by-step reasoning \citep{hsieh2023distilling,magister2023teaching}, with substantial gains reported in recent years. Task-agnostic distillation, in contrast, aims to replicate the teacher’s general instruction-following capability \citep{liu2026understanding,wang2023selfinstruct,mukherjee2023orca,agarwal2024onpolicy}. Its query distribution therefore needs to span a broad range of domains, typically requiring a substantially larger data-collection budget. In practice, task-specific  distillation represents a particularly relevant threat for API-based model extraction, as a distiller can concentrate its query budget on a high-value capability. This threat model is also supported by recent industry evidence: Anthropic\footnote{https://www.anthropic.com/news/detecting-and-preventing-distillation-attacks} reported large-scale multi-account distillation campaigns targeting a small number of high-value capabilities, including reasoning, tool use, and coding. This real-world evidence motivates our focus on the task-specific  distillation regime.

\noindent\textbf{Anti-distillation by perturbing distribution.}
A natural defense strategy is to perturb the model’s outputs so as to prevent the distiller from reliably recovering the targeted capability from the collected data.
\citet{orekondy2019prediction} perturbed the returned posterior probabilities to maximally distort attacker's gradient signal.
\citet{kariyappa2020defending} proposed to return incorrect predictions for out-of-distribution queries while preserving correct outputs for in-distribution inputs.
\citet{ma2021undistillable} fine-tuned a teacher model that encourages the distribution over incorrect labels to be uniform.
More recently, \citet{savani2025antidistillation} modified the token-level distribution, which steers sampling toward tokens that increase the student's loss on downstream tasks. However, these approaches inherently introduced bias, which can degrade the output quality perceived by benign users.

\noindent\textbf{Anti-distillation by adding watermark.}
Another line of work embeds identifiable signals into the model's outputs so that distillation can be detected post hoc.
\citet{zhao2023protecting} proposed GINSEW, which injects a secret sinusoidal signal into the sampling process.
\citet{sander2024watermarking} showed that text watermarks are ``radioactive'': when a student model is fine-tuned on watermarked text, residuals of the watermark signal persist in the student and can be statistically detected even when only a small fraction of the training data is watermarked.
\citet{xu2026antidistillation} proposed Antidistillation Fingerprinting, which injects a fingerprint penalty into the generation process. While watermarking provides a mechanism for attribution and legal recourse, it does not prevent the distillation itself, and most schemes also alter the output distribution.

\noindent\textbf{Anti-distillation with session-dynamic defense.}
Rather than perturbing every response, session-dynamic methods monitor user query patterns and activate defenses only when suspicious behavior is detected.
\citet{juuti2019prada} proposed PRADA, a detection-based defense that monitors the distribution of consecutive API queries and raises an alarm when this distribution deviates from benign user behavior.
\citet{chen2025queen} proposed QUEEN, which continuously monitors the sensitivity of incoming queries and begins perturbing output logits once the query pattern exceeds a predefined threshold.
\citet{mei2025defense} introduced D-ADD, which identifies malicious queries by analyzing per-account distribution discrepancy.
However, these approaches can be circumvented by a multi-account adversary who distributes queries across many accounts, each individually appearing benign.

\section{Method}
In this section, we formalize the multi-account distillation threat, present a warm-up algorithm \modelnamesimple{} for unconditional generation, and extend it to the full conditional setting \modelname{} via a semantic-bucket noise coupling mechanism.
\subsection{Problem Definition, Notations and Motivation}\label{sec:setup}
We denote by $\mathcal{X}$ the data space (e.g., image, text). Given any query $q\in\mathcal{Q}$ (e.g., a text prompt), the service provider possesses a private generative model that can produce samples from the conditional distribution $P_{\mathcal{X}}(\cdot \mid q)$. We regard this conditional distribution as the teacher model. Note that sampling from $P_{\mathcal{X}}(\cdot \mid q)$ can often be reparameterized as first drawing a noise variable
$\epsilon$ (or a collection of independent noise variables) from a simple distribution, and then producing the output via a deterministic mapping $G(q, \epsilon)$. Appendix~\ref{app:unified} provides detailed constructions for diffusion-based image generation and autoregressive text generation.
For convenience and ease of presentation, throughout the remainder of the paper, we represent the sampling process by $G(q, \epsilon)$ instead of $P_{\mathcal{X}}(\cdot \mid q)$. Accordingly, we write $x=G(q, \epsilon)$ to denote a sample generated by the model $G$, regardless of whether $x$ is an image or a sequence of text, and regardless of whether $\epsilon$ is a single noise variable, as in typical image generation, or a collection of noise variables, as in language generation.

\textbf{Definition of the distillation adversary.} A single-account distiller can be readily detected through behavioral patterns and subsequently restricted if it needs to collect a large amount of data within a short period for fine-tuning. We therefore focus on a distributed distiller who controls $K$ accounts, indexed by $k \in \{1,\dots,K\}$, to distill from the teacher model $G$. Given a limited collection period, we assume that each account issues $T$ queries at most. Let $q_{t,k}$ denote the $t$-th query submitted by account $k$,
 with $x_{t,k}$ denoting the corresponding response induced by noise $\epsilon_{t,k}$. We denote by $D_{\mathrm{distill}}$ the collection of query-response pairs obtained from these accounts. The distiller then fine-tunes a student model, parameterized by $\theta \in \Theta$, using an empirical risk of the general form
\begin{equation}\textstyle
 L_{\mathrm{standard}}(\theta)=\frac{1}{|D_{\mathrm{distill}}|}
    \sum_{(q,x)\in D_{\mathrm{distill}}}
    \ell(\theta; q, x)=\frac{1}{KT}\sum_{k=1}^{K}\sum_{t=1}^{T}\ell(\theta;q_{t,k}, x_{t,k}),
\end{equation}
where $\ell(\theta; q, x)$ denotes the per-instance training objective. In supervised fine-tuning, $\ell$ may take the form of a negative log-likelihood loss for language models or a score-matching loss for diffusion models. In more general reinforcement- or preference-based fine-tuning pipelines, $\ell$ may represent any surrogate objective computed from the observed query-response pairs.

\textbf{Definition of lossless benign-user experience.} Ideally, a benign user accesses the model through a single account to obtain responses. We call the experience lossless if a user receives the same experience under the defense mechanism as they would without any defense. To formalize this concept, we decompose losslessness into two key properties: distribution preservation and independence.

\textbullet\quad  The distribution-preservation property holds if, for any individual account—-regardless of whether it is benign or controlled by an adversary—-each response follows the original teacher distribution conditioned on the submitted query, thereby preserving per-query response fidelity.

\textbullet\quad  The independence property holds if the responses observed across accesses from the same account remain independent, thereby preserving the original repeated-sampling behavior within each account. 

Existing defenses primarily modify $G$ to introduce bias into the collected data, thereby breaking the distribution-preservation property. Rather than modifying $G$, we control the realization of the noise $\epsilon$ so as to preserve within-account independence while introducing cross-account dependence.

\subsection{A Warm-up Case: Anti-distillation Sampling in Unconditional Generation }\label{sec:unconditional_generation}

For sake of clarity, we begin with the simplest setting in which the user query $q$ is omitted. This setting is known as unconditional generation and has been widely studied in the generative modeling literature \citep{goodfellow2014generative,ho2020denoising,radford2018improving}. In this case, the generation mapping $G(q, \epsilon)$ and distillation loss $\ell(\theta; q, x)$ reduce to $G(\epsilon)$ and $\ell(\theta; x)$, respectively.

To achieve the goal, we leverage the coupling method, a classical tool in stochastic-process theory \citep{andersson2012stochastic,lindvall2002lectures}. Our algorithm \modelnamesimple{} associates the noise $\epsilon$ with the access count of each account. Specifically, let an injective mapping $\mathrm{SG}:\mathbb{N}\to\mathbb{N}$
denote a private seed-generation function maintained by the service
provider, which assigns the seed $s=\mathrm{SG}(t)$ to the $t$-th
access of any account. A standard seeded pseudorandom sampling routine is then used to draw $\epsilon$ from its target distribution. In modern sampling libraries, this can be implemented as $\epsilon=\mathsf{Expand}(F_\kappa(s))$, where $F_\kappa$ is a keyed pseudorandom generator with a secret key $\kappa$ sampled once at
deployment, and $\mathsf{Expand}$ maps the resulting pseudorandom bit sequence to the target noise distribution (e.g., a Gaussian distribution). The output $x=G(\epsilon)$ is then returned to the account.

We first establish that \modelnamesimple{} is lossless for each individual account. Under the standard pseudorandomness assumption, distinct seeds produce independent pseudorandom realizations from the same target distribution. Since $G$ remains unchanged and $\epsilon$ follows its original target distribution, the response $x$ is distributed exactly as a sample from
the teacher model, and thus the distribution-preservation property holds. Moreover, because $\mathrm{SG}$ is injective, distinct access indices $t\neq t'$ are mapped to distinct seeds, so the corresponding noise variables within any fixed account are independent across accesses. Therefore, the independence property also holds. The result follows immediately by combining the two properties.

\modelnamesimple{} can be viewed as a common-random-number coupling \citep{thorisson2000coupling} of the generative distributions across accounts. We next show how this strategy hinders distillation. While each individual account receives independent responses across accesses through distinct seeds, accesses from different accounts at the same access index are assigned the same seed. Thus, across the $K$ accounts controlled by the distiller, the $t$-th access uses the same noise, i.e., $\epsilon_{t,1}=\cdots=\epsilon_{t,K}$, and consequently produces the same output, $x_{t,1}=\cdots=x_{t,K}$. If the distiller uses these data for fine-tuning, the objective (1) reduces to
\begin{equation}\textstyle
 L_{\modelnamemath}(\theta)=\frac{1}{KT}\sum_{k=1}^{K}\sum_{t=1}^{T}\ell(\theta;x_{t,k})=\frac{1}{T}\sum_{t=1}^{T}\ell(\theta;x_{t,1}),
\end{equation}
To theoretically formalize how this mechanism hinders the distiller, we analyze the learning process through the lens of uniform convergence. Let
$\mathcal{F}:=\{x \mapsto \ell(\theta; x) : \theta \in \Theta\}$
be the induced loss class, and assume that \(\ell(\theta; x) \in [0,B]\) for all \(\theta \in \Theta\) and \(x\). We define the population risk as
    $\mathcal{E}(\theta)
    :=
    \mathbb{E}_{x \sim P_{\mathcal{X}}}[\ell(\theta; x)]$. The following proposition gives a uniform convergence bound between the empirical risk and the population risk.

\refstepcounter{definition}\label{prop:collapse}
\begin{tcolorbox}[theoremboxstyle, title={\textbf{Proposition \thedefinition\quad Effective Sample Size Collapse}}]
Let \(\mathfrak{R}_n(\mathcal{F})\) denote the Rademacher complexity of \(\mathcal{F}\) with sample size \(n\). For any \(\delta > 0\), with probability at least \(1-\delta\), the following hold:

\smallskip
1. For standard i.i.d.\ sampled data in (1), we have
\begin{equation}\textstyle
    \sup_{\theta \in \Theta}
    \left|L_{\mathrm{standard}}(\theta)-\mathcal{E}(\theta)\right|
    \le
    2\mathfrak{R}_{{\color{blue}K}{\color{red}T}}(\mathcal{F})
    +
    B\sqrt{\frac{\log(1/\delta)}{2{\color{blue}K}{\color{red}T}}}.
\end{equation}

2. For \modelnamesimple{} sampled data in (2), we have
\begin{equation}\textstyle
    \sup_{\theta \in \Theta}
    \left|L_{\modelnamemath}(\theta)-\mathcal{E}(\theta)\right|
    \le
    2\mathfrak{R}_{\color{red}T}(\mathcal{F})
    +
    B\sqrt{\frac{\log(1/\delta)}{2{\color{red}T}}}.
\end{equation}
\end{tcolorbox}

The above result follows directly from the standard uniform convergence theorem, and the detailed proof is provided in Appendix~\ref{app:proof-collapse}. Although it states only upper bounds, classical results show that Rademacher-based bounds of this form are tight up to constants \citep{bartlett2002rademacher, koltchinskii2006local, wainwright2019high}, so the gap between the two rates reflects a genuine statistical separation. To interpret the bound, we adopt the canonical scaling $\mathfrak{R}_n(\mathcal{F}) \le C/\sqrt{n}$, which holds for many hypothesis classes, including linear predictors, multilayer perceptrons, and Transformers under suitable norm constraints \citep{neyshabur2015norm,golowich2018size,trauger2024sequence}. Under this scaling, the generalization gap under \modelnamesimple{} is worse than that under standard i.i.d.\ sampling by a factor of $\sqrt{K}$, demonstrating a substantial reduction in learning effectiveness. Under \modelnamesimple{}, the generalization error depends only on the per-account query budget $T$. The distiller can improve learning efficiency only by substantially increasing $T$. This either raises the risk of detection and restriction due to a higher query rate per account, or requires a much longer distillation period, thereby reducing the benefit of rapid model catch-up. 

A drawback of \modelnamesimple{} is its limited diversity, which arises from the use of extreme coupling. A natural refinement is to use noise decomposition. For example, when the required noise is Gaussian, we can decompose it into a cross-account-coupled component and an independent component (see Experiment 4.1). In this way, within-account independence is preserved, cross-account dependency is maintained, and overall output diversity is improved. The corresponding uniform convergence analysis \citep{yu1994rates, mohri2008rademacher} has to be carried out case by case, depending on the specific randomization and the dependency structure. However, the general conclusion remains the same: introducing dependency degrades learning efficiency and makes generalization more difficult.

\refstepcounter{algorithm}\label{alg:LADS-simple}
\begin{algorithmbox}{Algorithm \thealgorithm\quad \modelnamesimple{}}
\begin{algorithmic}[1]
\Require Generative model $G$, private seed generator $\mathrm{SG}:\mathbb{N}\to\mathbb{N}$, keyed pseudorandom generator $F_\kappa$ with secret key $\kappa$, per-account access counters $\{t_u\}$ initialized to $0$
\Function{ServeRequest}{$u$}
    \State $t_u \gets t_u+1$ \Comment{Update the access count of account $u$}
    \State Seed $s \gets \mathrm{SG}(t_u)$
    \State Draw noise $\epsilon \gets \mathsf{Expand}(F_\kappa(s))$
    \State Generate output $x \gets G(\epsilon)$
    \State \Return $x$
\EndFunction
\end{algorithmic}
\end{algorithmbox}

\subsection{Anti-distillation Sampling for Conditional Generation}\label{sec:conditional_lads}

\modelnamesimple{} introduces dependence across accounts through noise coupling. We now extend this idea to the more realistic conditional-generation setting, where users may submit different queries and receive query-conditioned responses. In this setting, directly applying \modelnamesimple{} may induce only weak cross-account dependence: Although two accounts may share the same noise at a given access index, their outputs are given by $G(q,\epsilon)$ and $G(q', \epsilon)$; when $q$ and $q'$ differ substantially, the effect of the shared noise may be dominated by the difference between the queries. 

To overcome this limitation, we extend \modelnamesimple{} to incorporate query semantics into the coupling process, leading to our conditional-anti-distillation method, \modelname{}. We assume that the whole query space $\mathcal{Q}$ can be partitioned into $p$ semantic buckets, as specified by a private mapping $h: \mathcal{Q} \to [p] := \{1, 2, \dots, p\}$. For each user, we maintain a bucket-count array $a \in \mathbb{N}^p$, where $a[i]$ records the number of previous queries from this user that have fallen into
bucket $i$. Initially, all entries in $a$ are set to zero. When a user submits a query $q$, the provider first identifies the semantic bucket
$i = h(q)$ and updates the corresponding counter for this specific user:
\[\textstyle
a[i] \gets a[i] + 1.
\]
The random seed $s$ is then determined by
\[\textstyle
s = \mathrm{SG}(i, a[i]),
\]
where the seed generator $\mathrm{SG}: [p] \times \mathbb{N} \to \mathbb{N}$ is a private injective mapping. Similar to \modelnamesimple{}, this seed $s$ is
used to draw $\epsilon$. The final output is then generated as $x = G(q, \epsilon)$. The full procedure is summarized in
Algorithm~\ref{alg:LADS-conditional}. \modelname{} ties the generation noise to both the semantics of the query and the user’s access count within the corresponding semantic bucket. It is straightforward to verify that both the distribution-preservation and independence properties continue to hold. Therefore, \modelname{} remains lossless for each individual account.

\refstepcounter{algorithm}\label{alg:LADS-conditional}
\begin{algorithmbox}{Algorithm \thealgorithm\quad \modelname{}}
\begin{algorithmic}[1]
\Require Generative model $G$, private clustering function $h: \mathcal{Q} \to [p]$, private seed generator $\mathrm{SG}: [p] \times \mathbb{N} \to \mathbb{N}$, keyed pseudorandom generator $F_\kappa$ with secret key $\kappa$, per-account bucket-count arrays $\{a_u\in\mathbb{N}^p\}$ initialized to $0$
\Function{ServeRequest}{$u, q$}
    \State $i \gets h(q)$ \Comment{Identify the semantic bucket of query $q$}
    \State $a_u[i]\gets a_u[i]+1$ \Comment{Update the bucket count of account $u$}
    \State Seed $s \gets \mathrm{SG}(i, a_u[i])$
    \State Draw noise $\epsilon \gets \mathsf{Expand}(F_\kappa(s))$
    \State Generate output $x \gets G(q, \epsilon)$
    \State \Return $x$
\EndFunction
\end{algorithmic}
\end{algorithmbox}

As discussed in the related work, task-specific  distillation represents an important practical threat. We next show that, under reasonable conditions, \modelname{} can effectively hinder distillation in this setting. Let $\mathcal{Q}_{\mathrm{task}}\subseteq\mathcal{Q}$ denote the query space associated with the task, let $P_{\mathcal{Q}_{\mathrm{task}}}(\cdot)$ be the corresponding query distribution, and let $d_{\mathcal{Q}}(q,q')$ be a semantic metric on $\mathcal{Q}$, which in practice can be instantiated by measuring the
distance in the embedding space. Because the distiller targets a specific downstream capability, its queries are often semantically concentrated, in the sense that they occupy only a few localized subregions of the query space rather than being broadly distributed. Denote $H(\mathcal{Q}_{\mathrm{task}})=\{h(q)\mid q\in\mathcal{Q}_{\mathrm{task}}\}$ for the set of semantic buckets triggered by the task. We then assume that $|H(\mathcal{Q}_{\mathrm{task}})|=N\ll p$.

We assume that each semantic bucket admits a center under $d_{\mathcal{Q}}$, and that the private mapping $h$ assigns each query to its nearest center. Without loss of generality, let $c_1,\dots,c_N$ denote the centers of the N triggered buckets, and let $R$ denote the maximum distance between any query $q\in\mathcal{Q}_{\mathrm{task}}$ and its assigned
center. Thus, $R$ measures how tightly the task queries are concentrated within their corresponding semantic buckets. We define the natural conditional population risk as:
\begin{equation}\textstyle
    \mathcal{E}(\theta) := \mathbb{E}_{q \sim P_{\mathcal{Q}_{\text{task}}}}\mathbb{E}_{x \sim P_{\mathcal{X}}(\cdot \mid q)}[\ell(\theta; q, x)].
\end{equation}
The distiller's empirical risk is
 $L_{\text{LADS}}(\theta) = \frac{1}{KT}\sum_{k=1}^K\sum_{t=1}^T \ell(\theta; q_{t,k}, x_{t,k})$, following previous notations. Correspondingly, we define the loss function class for conditional generation as $\mathcal{F}:=\{(q,x)\mapsto\ell(\theta;q,x):\theta\in \Theta\}$ and establish the following uniform convergence bound.

\refstepcounter{definition}\label{thm:conditional_main}
\begin{tcolorbox}[theoremboxstyle, title={\textbf{Theorem \thedefinition\quad Effective Sample Size Collapse for \modelname{}}}]
Assume $G(q, \epsilon)$ is $L_G$-Lipschitz in $q$ under the metric \(d_{\mathcal{Q}}\), and the loss $\ell$ is $L_\ell$-Lipschitz with respect to the input pair $(q,x)$ under their joint metric, and bounded by $B$. For any $\delta \in (0, 1)$, with probability at least $1 - 2\delta$, we have
\begin{equation}\small\textstyle
    \textstyle\sup_{\theta \in \Theta} |L_{\text{LADS}}(\theta) - \mathcal{E}(\theta)| \le
    \underbrace{2\mathfrak{R}_{NT}(\mathcal{F}) + \textstyle B\sqrt{\frac{\log(1/\delta)}{2NT}}}_{\text{Estimation Error w.r.t.\ Surrogate Target}}
    + \underbrace{\vphantom{\sqrt{\frac{\log(1/\delta)}{2n_{\mathrm{eff}}}}} 2L_\ell(1+L_G)R + \Delta_{\mathrm{align}}}_{\text{Systematic Bias}},
\end{equation}
where $\mathfrak{R}_{NT}(\mathcal{F}) = \mathcal{O}(1/\sqrt{NT})$ is the corresponding Rademacher complexity, and $\Delta_{\mathrm{align}} :=\textstyle  BN\sqrt{\frac{\log(2N/\delta)}{2KT}}$ captures the statistical fluctuation of cluster weights.
\end{tcolorbox}

The proof is provided in the appendix~\ref{app:conditional_lads_refined}.  Theorem~\ref{thm:conditional_main}
relies on the Lipschitz regularity of $G$ and $\ell$, an assumption commonly used to make theoretical analyses of neural networks tractable~\citep{edelman2022inductive,li2023transformers,castin2024smooth}. Based on the theorem above, we can see that for \modelname{}, the generalization gap shrinks at rate $\mathcal{O}(1/\sqrt{NT})$ rather than the i.i.d.\ rate $\mathcal{O}(1/\sqrt{KT})$, so a substantially larger gap persists whenever the number of affected clusters is much smaller than the number of controlled accounts, namely $N\ll K$. Moreover, both the number and quality of the semantic clusters affect the performance of the algorithm. When $p$ is large and the clustering is fine-grained, queries within each bucket are more semantically concentrated, leading to a smaller $R$ and potentially smaller values of $L_G$ and $L_\ell$, at the cost of increased storage overhead. In contrast, when $p$ is small or the clustering quality is poor, queries within the same bucket are less semantically coherent, resulting in a larger $R$ and potentially larger values of $L_G$ and $L_\ell$. This leads to a looser generalization bound and weaker control over the student model’s generalization performance. 

\textbf{Practical implementation for large-scale system.}
Industrial-scale deployment of LADS requires scalable construction of the bucket assignments $h$ and maintenance of per-user count arrays, both of which can leverage well-established infrastructure with manageable overhead. First, the bucket assignments $h$ can be instantiated by embedding each query into a dense vector and assigning it to a bucket via locality-sensitive hashing or, alternatively, through a nearest-center lookup in the embedding space. Mature and highly optimized systems are readily available, including FAISS \citep{johnson2021billion,douze2026faiss} and ScaNN \citep{guo2020accelerating}, which support billion-scale dense retrieval with millisecond-level query latency. Second, although we write the bucket counter as a dense array $a\in\mathbb{N}^{p}$ for notational convenience, it can be implemented sparsely. Since each account typically touches only a small subset of the $p$ buckets, the provider only needs to maintain counters for those visited buckets. This requires only lightweight state access and keeps the memory overhead manageable even when the global bucket space is large.

\textbf{Potential adversarial strategy and task-agnostic distillation.} 
Our current analysis primarily targets task-specific distillation, where the adversary’s queries concentrate in a relatively small number of semantic regions. LADS can also be applied to task-agnostic distillation, but its effectiveness may decrease as the number $N$ of activated semantic buckets grows, thereby reducing the amount of cross-account coupling. This reveals a natural trade-off in semantic granularity: finer clustering yields tighter semantic coherence but activates more buckets, whereas coarser clustering increases coupling coverage at the cost of weaker dependence among outputs generated from queries within the same bucket. LADS may also face an adaptive distiller that strategically modifies its behavior to circumvent the defense. For example, the distiller may attempt to reverse-engineer the bucket assignment and then either concentrate semantically similar queries within a single account or adversarially rewrite them so that they are mapped to different buckets. Both strategies can increase the independence of the collected samples. Although such reverse engineering is highly nontrivial, its difficulty alone does not constitute a formal security guarantee. This motivates an empirical evaluation of LADS under imperfect or noisy semantic mappings, which we present in the experimental section.

\section{Experiment}
In this section, we evaluate \modelname{} across two scenarios and two representative generative paradigms: diffusion-based image generative models and autoregressive language models. Both experimental results support the effectiveness of our proposed algorithm.

\subsection{Image Generation}
\label{subsec:exp_image}

We conduct experiments on class-conditional image generation using the EDM2 models~\citep{karras2024analyzing}. EDM2 is a family of pre-trained diffusion models on the ImageNet dataset. For all experiments, the checkpoints with the lowest validation loss are reported.

\textbf{Simulating multi-account distillation.}
In our experiment, we employ two pretrained teacher models, EDM2-Large (777M parameters) and EDM2-XS (125M parameters), and we use EDM2-XXS (31M parameters) as the student model. They share the same UNet backbone and differ only in the base channel width. For both teacher-side data generation and student-side evaluation, we use a 32-step second-order Heun ODE sampler.  To emulate the threat model, we set the total query budget to be 1M, and set the number of distiller's accounts $K=50$. Therefore, each of the accounts issues $T=20k$ queries. The student model is then trained on the collected dataset, with all training/inference hyperparameters following the original EDM2 settings.

\textbf{Simulating noisy semantic mapping.}
In this task, we represent each query $q$ as a class label. To simulate a more realistic setting, we construct non-overlapping buckets, each containing four randomly selected class labels. This mimics practical scenarios where semantically distinct queries may be assigned to the same bucket because of clustering errors. Furthermore, for each submitted query, we assign it to the true bucket with probability $\beta$, and to a randomly selected bucket with probability $1-\beta$, where $\beta$ controls the noise level.

\textbf{\modelname{} implementation and distillation training.}
In EDM2 models, $\epsilon$ follows a standard Gaussian distribution. To quantify the strength of cross-account noise coupling and its impact on distillation performance, we decompose $\epsilon$ into a \emph{bucket-count-shared} component and a \emph{per-request independent} component. Concretely, let $\mathrm{GRNG}(s)$ denote a Gaussian Random Number Generator that outputs a sample from $\mathcal{N}(\mathbf{0}, \mathbf{I})$ when initialized with seed $s$, and let $\mathrm{GRNG}()$ denote the same generator with a randomly chosen seed. $\mathrm{SG}$ is a random number list and fixed during our experiment. For any user submitting query that falls into the $i$-th bucket under the noisy mapping, the provider computes the seed $s = \mathrm{SG}(i, a[i])$ and constructs the noise $\epsilon$ according to:
\begin{equation}\label{eq:noise_coupling}\textstyle
    \epsilon = \sqrt{\alpha} \cdot \mathrm{GRNG}(s) + \sqrt{1 - \alpha} \cdot \mathrm{GRNG}(),
\end{equation}
where $\alpha \in [0, 1]$ is the mixing coefficient. It is straightforward to verify that \(\epsilon\) still follows \(\mathcal{N}(\mathbf{0}, \mathbf{I})\), thereby ensuring that generation remains lossless for benign users. The parameter $\alpha$ directly controls the strength of cross-account coupling: when $\alpha = 0$, all noise variables are independent, and the resulting sampling procedure reduces to standard i.i.d.\ sampling regardless of the value of $\beta$; $\alpha = 1$ recovers the extreme coupling regime of \modelname{}. 

\textbf{Evaluation metrics.} We evaluate the distilled student model under different configurations using two classes of metrics.
The first class is the Fr\'{e}chet Inception Distance (FID)~\citep{heusel2017gans}.
We consider two comparisons: FID (to real), which calculates the FID score between the student-generated images and the real ImageNet images, and FID (to teacher), which measures how faithfully the student recovers the teacher's output distribution.
The second class is the loss value, which more directly reflects the approximation and generalization of the learned models. To evaluate the quality experienced by a benign user, we fix a user account, generate samples independently, and compute the FID between the generated images and the real ones.

\textbf{Results.} Detailed results are presented in Table~\ref{tab:image_results}. We observe that the coupling strength $\alpha$ affects the performance of the student model in terms of both FID and validation loss. For the EDM2-Large teacher, FID (to real) rises from $8.50$ at $\alpha=0$ to $29.8$ at $\alpha=1.0$, and FID (to teacher) grows from $4.53$ to $26.95$, with a similar trend for the EDM2-XS teacher. Moreover, even under an imperfect semantic mapping ($\beta<1$), the student model still exhibits degraded performance, indicating that LADS may remain effective in realistic settings. The validation loss also increases monotonically with $\alpha$, indicating a growing failure to generalize to the true data distribution. As for the training loss, note that checkpoints are selected based on validation loss, so the total number of training steps differs across experiments. At $\alpha=0.0$, the number of training steps is significantly larger than in the other settings, suggesting that the model has not yet overfit. In contrast, at $\alpha=1.0$, the severely reduced data diversity causes the training loss to decrease much more rapidly. For benign users, the FID remains similar to \citet{karras2024analyzing}, indicating that they receive high-quality samples.
\begin{table}[t]
  \centering
  \small
  \setlength{\tabcolsep}{5.5pt}
  \renewcommand{\arraystretch}{1.08}
    \begin{tabular}{lllccccc}
      \toprule
      \multirow{2}{*}{Teacher} & \multirow{2}{*}{$\alpha$} & \multirow{2}{*}{$\beta$} & \multicolumn{2}{c}{Student FID} & \multicolumn{2}{c}{Student Loss} & \multirow{2}{*}{Benign User FID} \\
      \cmidrule(lr){4-5} \cmidrule(lr){6-7}
      & & & To real & To teacher & Train & Valid & \\
      \midrule
      \multirow{4}{*}{EDM2-Large}
        & 0.0 & 0.0 & {\color{blue}8.50}  & {\color{blue}4.53}  & {\color{blue}0.301} & {\color{blue}0.340} & \multirow{4}{*}{2.06} \\
        & 0.7 & 0.7 & 9.20               & 5.20               & 0.305                & 0.346                & \\
        & 0.9 & 0.9 & 21.09              & 16.32              & 0.317                & 0.366                & \\
        & 1.0 & 1.0 & {\color{red}29.8}  & {\color{red}26.95} & {\color{red}0.130}  & {\color{red}0.848}  & \\
      \midrule
      \multirow{4}{*}{EDM2-XS}
        & 0.0 & 0.0 & {\color{blue}9.39}  & {\color{blue}2.80}  & {\color{blue}0.309} & {\color{blue}0.345} & \multirow{4}{*}{3.53} \\
        & 0.7 & 0.7 & 9.75               & 3.13               & 0.312                & 0.349                & \\
        & 0.9 & 0.9 & 22.67               & 13.61              & 0.322               & 0.370                & \\
        & 1.0 & 1.0 & {\color{red}25.94}  & {\color{red}29.76} & {\color{red}0.119}  & {\color{red}0.823}  & \\
      \bottomrule
    \end{tabular}%
  \vspace{4pt}
  \caption{Evaluation results of student models. When $\alpha=0$ (highlighted in blue), corresponding to the standard i.i.d.\ setting, the student model performs well in terms of both FID and loss. As $\alpha$ and $\beta$ both increase toward $1$ (highlighted in red), distillation performance becomes progressively worse, indicating that the proposed algorithm is effective. }
  \label{tab:image_results}
  \vspace{-15pt}
\end{table}

\subsection{Language Model}
\label{subsec:exp_language}

Mathematical reasoning and code generation are among the most representative and widely studied distillation targets, as demonstrated by recent works such as DART-Math~\citep{tong2024dart}, MathFusion~\citep{pei2025mathfusion}, WizardCoder~\citep{luo2023wizardcoder}, and Magicoder~\citep{wei2023magicoder}. Following this line of work, we adopt a similar distillation pipeline: a frontier-scale teacher generates responses for a curated set of prompts, and a smaller student is then fine-tuned on these responses. We use Qwen3.5-397B-A17B-FP8~\citep{qwen3.5} as the teacher model. For mathematical reasoning, we consider Llama-3.1-8B~\citep{grattafiori2024llama} and DeepSeek-Math-7B-Base ~\citep{deepseek-math} as student models; for code generation, we use Qwen2.5-7B~\citep{qwen2025qwen25technicalreport} as the student model.


\textbf{LADS implementation and distillation training.}
 We use training prompts from MATH~\citep{hendrycks2021measuring} and GSM8K~\citep{cobbe2021training} for mathematical reasoning, and Code Alpaca~\citep{codealpaca} for code generation. For each base prompt, the teacher generates 9 semantically equivalent rephrasings. In the standard i.i.d.\ setting, the teacher independently rolls out all 10 variants. In the \modelname{} setting, we set the numbers of semantic buckets to 7.5k, 7.5k, and 20k for the three tasks, respectively. Each query is then mapped to a semantic bucket through nearest-neighbor search in the embedding space.  Same-bucket prompts are assigned the same decoding seed with probability 0.7 and different seeds otherwise, simulating a noisy semantic mapping. For teacher rollout, we sample with a maximum token length of 65,536, temperature 0.7, top-$p$ of 0.8, top-$k$ of 20, and a presence penalty of 1.5. The thinking mode of the teacher is disabled during rollout. All student models are fine-tuned for 2 epochs with a learning rate of $2 \times 10^{-5}$, cosine learning rate schedule with a warmup ratio of 0.01, weight decay of 0.1. For evaluation, we report the accuracy on GSM8K and MATH test set, and pass@1 on HumanEval~\citep{chen2021evaluating} using the official unit tests.

\textbf{Results.}
Slightly different from prior works, we adopt a zero-shot evaluation protocol for all models to ensure a fair comparison across different sampling settings, which results in lower baseline numbers than those reported in the original works. As shown in Table~\ref{tab:language_results}, the standard i.i.d.\ sampling yields substantial gains over the pre-trained student across all models and benchmarks (e.g., +29.8 and +49.5 points on GSM8K).
In contrast, under \modelname{} sampling, the improvement is significantly reduced. Specifically, on GSM8K, the distillation gain drops from +29.8 to +11.2 for Llama-3.1-8B and from +49.5 to +35.1 for DeepSeek-Math-7B, corresponding to absolute reductions of 18.6 and 14.4 points, respectively, and a similar trend is observed on MATH and HumanEval. These results confirm that \modelname{} effectively limits the statistical benefit of multi-account distillation.

\begin{table}[t]
  \centering
  \small
  \setlength{\tabcolsep}{5.5pt}
  \renewcommand{\arraystretch}{1.08}
  \begin{tabular}{llccc}
    \toprule
    \multirow{2}{*}{Benchmark} & \multirow{2}{*}{Model} & No SFT & Standard i.i.d Sampling & LADS Sampling\\
    & &(the pre-trained student) & (SFT student) & (SFT student) \\
    \midrule
    \multirow{2}{*}{GSM8K}
      & Llama-3.1-8B      & 43.59\% & 73.39\% (+{\color{blue}29.8}) & 62.72\% (+{\color{red}19.1}) \\
      & DeepSeek-Math-7B  & 23.70\% & 73.16\% (+{\color{blue}49.5}) & 60.13\% (+{\color{red}36.4}) \\
    \midrule
    \multirow{2}{*}{MATH}
      & Llama-3.1-8B      & 13.54\% & 28.18\% (+{\color{blue}14.6}) & 22.47\% (+{\color{red}8.9}) \\
      & DeepSeek-Math-7B  &  6.20\% & 32.66\% (+{\color{blue}26.5}) & 27.57\% (+{\color{red}21.4}) \\
    \midrule
    HumanEval
      & Qwen2.5-7B        & 56.70\% & 71.95\% (+{\color{blue}15.3}) & 65.20\% (+{\color{red}8.5}) \\
    \bottomrule
  \end{tabular}
  \vspace{4pt}
  \caption{Evaluation results of student models fine-tuned on samples collected under different sampling strategies of the teacher model. When the teacher uses standard i.i.d.\ sampling, student performance (highlighted in blue) improves significantly after fine-tuning on these samples. In contrast, the performance gains are substantially reduced under LADS sampling (highlighted in red).}
  \label{tab:language_results}
  \vspace{-15pt}
\end{table}

\section{Conclusion and Future Work}
We proposed Lossless Anti-Distillation Sampling (\modelname{}), a principled mechanism for mitigating multi-account distillation without compromising benign-user experience. Unlike prior defenses that rely on output perturbation or detection, \modelname{} operates at the level of randomness, preserving exact sampling fidelity for individual users while introducing structured dependence across accounts. Our theoretical analysis shows that this coupling mechanism fundamentally limits the statistical efficiency of distillation. Extensive experiments across both image and language domains demonstrate that \modelname{} weakens distillation performance while leaving benign-user outputs unaffected. Looking forward, an important direction is to design more adaptive and fine-grained semantic grouping strategies that further tighten cross-account coupling under realistic workloads. It is also of interest to study the interaction between \modelname{} and more advanced adversarial strategies. In addition, extending the framework to broader generative paradigms and integrating it with complementary system-level protections may enable a more comprehensive defense against large-scale distillation.

\section*{Ethics Statement}
This work studies a server-side defense that a model provider can deploy against
multi-account distillation of its own model. It does not introduce a new attack,
and it does not require collecting, storing, or analyzing any additional user
content beyond the per-user, per-bucket access counters described in
Section~\ref{sec:conditional_lads}. By construction the mechanism is lossless:
the response distribution seen by any individual account is identical to that of
the undefended model, so no user experiences degraded output quality. The main
foreseeable side effect is that legitimate parties who intentionally query the
service from many accounts with semantically equivalent prompts (for example,
large-scale academic evaluation efforts) would observe reduced sample diversity;
providers deploying \modelname{} should therefore offer sanctioned,
non-coupled access paths for such use cases. All datasets and pre-trained models
used in our experiments are publicly available and used in accordance with their
licenses.

\section*{Reproducibility Statement}
All components needed to reproduce our results are specified in the paper.
The full mechanism is given as pseudocode in Algorithm~\ref{alg:LADS-simple}
(unconditional) and Algorithm~\ref{alg:LADS-conditional} (conditional), and the
noise-coupling construction used in the image experiments is stated explicitly
in Equation~\eqref{eq:noise_coupling}. Section~\ref{subsec:exp_image} lists the
teacher/student checkpoints, sampler settings, query budget, and account count
for the EDM2/ImageNet experiments, and Section~\ref{subsec:exp_language} lists
the prompt pools, decoding parameters, and fine-tuning hyperparameters for the
language experiments. Assumptions and complete proofs for
Proposition~\ref{prop:collapse} and Theorem~\ref{thm:conditional_main} are given
in Appendices~\ref{app:proof-collapse} and~\ref{app:conditional_lads_refined}.

\section*{AI Use Statement}
Generative AI tools were used in this work in the following ways. First, they
are part of the object of study: the teacher language model was used to
construct the prompt pool, rephrasing each seed prompt into nine
semantics-preserving variants, and to generate the responses that constitute
the distillation training data for all student models
(Section~\ref{subsec:exp_language}). These generated data are experimental
material rather than content of the paper, and their construction is described
in full in the corresponding sections. Second, a large language model was used
as a general-purpose writing aid for grammar, phrasing, and wording, and for
routine coding assistance in the experimental scripts.

We did not use generative AI tools to develop the theoretical model or
conceptual framework, to formulate the mathematical claims, to write any of the
proofs, to propose or refine the research hypotheses, to design the research
methodology, or to interpret the experimental results. All AI-assisted output
was reviewed by the authors: generated text was checked for accuracy and
originality, and generated code was verified for correctness against the
reported experimental setup. We take responsibility for the final content of
this work, including all text, claims, and artifacts produced with the aid of
generative AI.

\bibliographystyle{plainnat}
\bibliography{references}


\appendix
\addtocontents{toc}{\protect\setcounter{tocdepth}{2}}

\newpage
\renewcommand{\contentsname}{Appendix Contents}
{\hypersetup{linkcolor=black}\tableofcontents}

\newpage

\section{Additional Related Work}
\label{app:additional_related_work}

\subsection{Knowledge Distillation}
Knowledge distillation, introduced by \citet{hinton2015distilling} for image classification, transfers a teacher's predictive distribution to a smaller student via temperature-softened cross-entropy. Subsequent work extended the framework along several axes: \citet{romero2014fitnets} matched intermediate-layer representations through hint losses; \citet{furlanello2018born} showed that distilling a model into an architecturally identical student (so-called \emph{self-distillation}) can improve generalization; and \citet{tian2020contrastive} proposed contrastive losses to capture richer structural information beyond logits. Distillation has also been adapted to language and sequence-to-sequence settings, notably via DistilBERT~\citep{sanh2019distilbert} and sequence-level knowledge distillation~\citep{kim2016sequence}. In modern large-language-model practice, distillation has become a dominant low-cost route to specialized capability: chain-of-thought traces sampled from frontier teacher models are routinely used as supervised fine-tuning data for smaller open models. Prominent examples include MiniLLM~\citep{gu2024minillm}, Distilling Step-by-Step~\citep{hsieh2023distilling}, and Teaching Small Language Models to Reason~\citep{magister2023teaching}, alongside domain-specialized pipelines such as DART-Math~\citep{tong2024dart}, MathFusion~\citep{pei2025mathfusion}, and DeepSeek-Math~\citep{deepseek-math} for mathematical reasoning. While academically productive, these pipelines also embody the threat model studied in this paper: any party with API access can replicate a substantial fraction of a frontier model's capability at a fraction of its original training cost.

\subsection{Model Extraction Attacks}
The general problem of harvesting a black-box model through its prediction API was first formalized by \citet{tramer2016stealing}, who showed that linear models, decision trees, and shallow neural networks can be extracted with high fidelity using only a moderate number of API queries. \citet{papernot2017practical} extended these ideas to surrogate-model attacks against deeper image classifiers, training a substitute model from API outputs and using it to craft transferable adversarial examples. \citet{jagielski2020high} systematically distinguished between \emph{accuracy} extraction (matching teacher predictions on natural inputs) and \emph{fidelity} extraction (matching teacher predictions even on adversarial inputs), establishing relationships between query budget and extraction quality. To remove the need for in-distribution queries, \citet{truong2021data} proposed Data-Free Model Extraction, which uses generative networks to synthesize informative queries, while \citet{pal2020activethief} exploited active learning to select the most informative public-domain queries. The same family of attacks has been demonstrated against language and translation systems: \citet{krishna2020thieves} stole BERT-style classifiers using random or out-of-distribution input queries, \citet{wallace2020imitation} attacked production black-box machine translation systems and proposed adversarial-example-based defenses, and more recently \citet{carlini2024stealing} recovered partial weights of production large language models through carefully designed query strategies. Collectively, these attacks form the technical foundation of the modern multi-account distillation threat: black-box query access alone is sufficient to approximate or partially reconstruct the teacher, motivating output-side and query-side defenses such as the one studied in this paper.

\subsection{Additional Defenses via Watermarking}
Beyond the three defense families discussed in Section~\ref{related_work}, a parallel line of work targets \emph{post hoc} ownership verification by embedding identifiable signals into the model itself. \citet{uchida2017embedding} embedded watermarks directly into model weights via a regularization term during training, while \citet{adi2018turning} repurposed backdoor triggers as ownership signatures that persist through fine-tuning. \citet{jia2021entangled} proposed entangled watermarks, designed to be inseparable from the model's primary task and thus robust to model extraction. These watermark-based methods complement \modelname{}: they support post-incident attribution, whereas \modelname{} actively prevents the adversary from obtaining a high-quality distilled model in the first place.

\section{Rademacher Complexity of the Softmax Linear Model}
\label{app:softmax-rademacher}

We verify the bound $\mathfrak{R}_n(\mathcal{F}) \le 2RW\sqrt{Y}/\sqrt{n}$ used in the main text. Consider the softmax linear model class
\[\textstyle
\mathcal{P}_\Theta
=
\Big\{\, p_\theta(y\mid x) \propto \exp(\langle \theta_y, x\rangle)\ :\ \theta=(\theta_1,\dots,\theta_Y),\ \|\theta\|_F \le W \,\Big\},
\qquad
\|\theta\|_F^2 = \sum_{y=1}^{Y}\|\theta_y\|_2^2,
\]
trained with the cross-entropy loss $\ell(\theta;x,y) = -\log p_\theta(y\mid x) = -\langle \theta_y,x\rangle + \log\sum_{y'=1}^{Y}\exp(\langle \theta_{y'},x\rangle)$ on inputs satisfying $\|x\|_2\le R$. Let $\mathcal{F} = \{(x,y)\mapsto \ell(\theta;x,y):\theta\in\Theta\}$ be the induced loss class, and let $h_\theta(x) = (\langle \theta_1,x\rangle,\dots,\langle \theta_Y,x\rangle)\in\mathbb{R}^{Y}$ denote the logit vector.

\textbf{Step 1: Lipschitz constant of the cross-entropy.}
Viewing the cross-entropy as a function of the logits $z\in\mathbb{R}^Y$ for a fixed label $y$, its gradient equals $\nabla_z \ell = \mathrm{softmax}(z) - e_y$, whose $\ell_2$-norm is bounded by
\[\textstyle
\|\mathrm{softmax}(z) - e_y\|_2
\;\le\;
\|\mathrm{softmax}(z)\|_2 + \|e_y\|_2
\;\le\;
1 + 1
\;=\; 2,
\]
and more tightly $\le \sqrt{2}$ since $\mathrm{softmax}(z)$ is a probability vector, so $\|\mathrm{softmax}(z)\|_2^2 \le \|\mathrm{softmax}(z)\|_1 = 1$. Hence $\ell$ is $L_\ell$-Lipschitz in $z$ with $L_\ell \le \sqrt{2}$.

\textbf{Step 2: Maurer's vector-contraction inequality.}
Let $\{\sigma_{iy}\}_{i\in[n],\,y\in[Y]}$ be i.i.d.\ Rademacher random variables. By \citet[Corollary~1]{maurer2016vector}, for any $L$-Lipschitz scalar function $\phi:\mathbb{R}^Y\to\mathbb{R}$ and vector-valued class $\mathcal{H}$,
\[\textstyle
\mathbb{E}\sup_{h\in\mathcal{H}}\sum_{i=1}^{n}\sigma_i\,\phi(h(x_i))
\;\le\;
\sqrt{2}\,L\,\mathbb{E}\sup_{h\in\mathcal{H}}\sum_{i=1}^{n}\sum_{y=1}^{Y}\sigma_{iy}\,h_y(x_i).
\]
Applying this with $\phi=\ell(\cdot;y)$ and $h_\theta = (\langle\theta_1,\cdot\rangle,\dots,\langle\theta_Y,\cdot\rangle)$ yields
\begin{equation}\textstyle
\label{eq:app-maurer}
\mathfrak{R}_n(\mathcal{F})
\;\le\;
\sqrt{2}\,L_\ell\,\cdot\,
\frac{1}{n}\,\mathbb{E}\sup_{\|\theta\|_F\le W}\sum_{i=1}^{n}\sum_{y=1}^{Y}\sigma_{iy}\langle \theta_y,x_i\rangle.
\end{equation}

\textbf{Step 3: Controlling the linear supremum.}
Define $v_y := \sum_{i=1}^{n}\sigma_{iy}\,x_i \in\mathbb{R}^d$. By Cauchy--Schwarz under the Frobenius constraint,
\[\textstyle
\sup_{\|\theta\|_F\le W}\sum_{y=1}^{Y}\langle \theta_y, v_y\rangle
\;=\;
W\,\sqrt{\sum_{y=1}^{Y}\|v_y\|_2^2}.
\]
Taking expectations and using Jensen's inequality,
\[\textstyle
\mathbb{E}\sqrt{\sum_{y=1}^{Y}\|v_y\|_2^2}
\;\le\;
\sqrt{\sum_{y=1}^{Y}\mathbb{E}\|v_y\|_2^2}
\;=\;
\sqrt{\sum_{y=1}^{Y}\sum_{i=1}^{n}\|x_i\|_2^2}
\;\le\;
\sqrt{Y\cdot n R^2}
\;=\;
R\sqrt{Y n},
\]
where we used that $\mathbb{E}[\sigma_{iy}\sigma_{jy}] = \mathbf{1}\{i=j\}$ to obtain $\mathbb{E}\|v_y\|_2^2 = \sum_i\|x_i\|_2^2$.

\textbf{Step 4: Combining.}
Substituting into \eqref{eq:app-maurer},
\[\textstyle
\mathfrak{R}_n(\mathcal{F})
\;\le\;
\sqrt{2}\,L_\ell\cdot\frac{W\cdot R\sqrt{Yn}}{n}
\;=\;
\frac{\sqrt{2}\,L_\ell\,R\,W\,\sqrt{Y}}{\sqrt{n}}
\;\le\;
\frac{2\,R\,W\,\sqrt{Y}}{\sqrt{n}},
\]
where the last inequality uses $L_\ell\le\sqrt{2}$. Setting $C := 2RW\sqrt{Y}$ recovers the scaling $\mathfrak{R}_n(\mathcal{F})\le C/\sqrt{n}$ claimed in the main text.\qed

\section{Unified Representation of Generative Models}\label{app:unified}

We denote by $\mathcal{X}$ the data space (e.g., image, text). Given any query $q\in\mathcal{Q}$ (e.g., a text prompt), the service provider possesses a private generative model that can produce samples from the conditional distribution $P_{\mathcal{X}}(\cdot \mid q)$. We regard this conditional distribution as the teacher model. Note that sampling from $P_{\mathcal{X}}(\cdot \mid q)$ can often be reparameterized as first drawing a noise variable
$\epsilon$ (or a collection of independent noise variables) from a simple distribution, and then producing the output via a deterministic mapping $G(q, \epsilon)$.

In diffusion-based image generative models \citep{sohl2015deep, ho2020denoising, song2020score, karras2022elucidatingdesignspacediffusionbased}, \(\epsilon\) is typically drawn from a standard multivariate Gaussian distribution, while the mapping function \(G\) is given by the corresponding ODE solver \citep{song2021scorebasedgenerativemodelingstochastic}. In language models, output tokens are generated autoregressively, with each token sampled from the categorical distribution over the vocabulary \(V\) induced by the softmax of the logits. Equivalently, this sampling procedure can be realized by first drawing a noise vector \(\epsilon \in \mathbb{R}^{|V|}\), whose entries are independently sampled from the standard Gumbel distribution \citep{gumbel1954statistical}, and then selecting the token index $\arg\max_{i \in V} \left( z_{i} + \epsilon_i \right)$ according to the Gumbel-max trick \citep{luce1959individual,yellott1977relationship,papandreou2011perturb,hazan2012partition,maddison2014sampling}, where \(z \in \mathbb{R}^{|V|}\) denotes the logit vector and $z_{i}$  is its $i$-th element.

\section{Proof of Proposition~\ref{prop:collapse}}
	\label{app:proof-collapse}

	We prove Proposition~\ref{prop:collapse} in a self-contained manner. Recall the
	unconditional-generation setting in Section~\ref{sec:unconditional_generation}, where the query \(q\) is omitted,
	the generation rule is written as \(x=G(\epsilon)\), and the distillation loss
	reduces to \(\ell(\theta;x)\). The induced loss class is
	\[\textstyle
	\mathcal F
	:=
	\{x\mapsto \ell(\theta;x):\theta\in\Theta\},
	\qquad
	0\le \ell(\theta;x)\le B
	\quad
	\text{for all }(\theta,x),
	\]
	and the population risk is
	\[\textstyle
	\mathcal{E}(\theta)
	:=
	\mathbb E_{x\sim P_{\mathcal X}}[\ell(\theta;x)].
	\]
	The proof consists of two steps. We first establish a generic uniform convergence
	bound for an arbitrary i.i.d.\ sample of size \(n\). We then apply this bound with
	\(n=KT\) for standard i.i.d.\ sampling and with \(n=T\) for \modelnamesimple{}.

	\textbf{Step 1: A self-contained uniform deviation bound.}
	Let \(Z_1,\dots,Z_n\stackrel{\mathrm{i.i.d.}}{\sim}P_{\mathcal X}\), and define
	\[\textstyle
	\Psi(Z_1,\dots,Z_n)
	:=
	\sup_{\theta\in\Theta}
	\left|
	\frac1n\sum_{i=1}^n \ell(\theta;Z_i)-\mathcal{E}(\theta)
	\right|.
	\]
	We claim that for every \(\delta\in(0,1)\), with probability at least \(1-\delta\),
	\begin{equation}\textstyle
		\label{eq:generic-uniform-bound}
		\Psi(Z_1,\dots,Z_n)
		\le
		2\mathfrak R_n(\mathcal F)
		+
		B\sqrt{\frac{\log(1/\delta)}{2n}}.
	\end{equation}

	We prove this claim in two parts.

	\smallskip
	\noindent
	\emph{Step 1a: Concentration around the mean.}
	Consider \(\Psi\) as a function of the sample \(Z_1,\dots,Z_n\). Fix an index
	\(j\in\{1,\dots,n\}\), and let
	\[\textstyle
	Z^{(j)}
	:=
	(Z_1,\dots,Z_{j-1},Z_j',Z_{j+1},\dots,Z_n)
	\]
	be the sample obtained by replacing \(Z_j\) with another value \(Z_j'\). For every
	fixed \(\theta\in\Theta\),
	\begin{align*}\textstyle
		&\textstyle \left|
		\left(
		\frac1n\sum_{i=1}^n \ell(\theta;Z_i)-\mathcal{E}(\theta)
		\right)
		-
		\left(
		\frac1n\sum_{i\ne j}\ell(\theta;Z_i)
		+
		\frac1n\ell(\theta;Z_j')
		-
		\mathcal{E}(\theta)
		\right)
		\right|  \\\textstyle
		&\textstyle \qquad =
		\frac1n
		\bigl|
		\ell(\theta;Z_j)-\ell(\theta;Z_j')
		\bigr|
		\le
		\frac{B}{n},
	\end{align*}
	because \(\ell(\theta;\cdot)\in[0,B]\). Moreover, for any two collections
	\(\{a_\theta\}_{\theta\in\Theta}\) and \(\{b_\theta\}_{\theta\in\Theta}\),
	\[\textstyle
	\left|
	\sup_{\theta\in\Theta} a_\theta
	-
	\sup_{\theta\in\Theta} b_\theta
	\right|
	\le
	\sup_{\theta\in\Theta}|a_\theta-b_\theta|.
	\]
	Applying this inequality to the absolute-deviation functions inside \(\Psi\), we
	obtain
	\[\textstyle
	\left|
	\Psi(Z_1,\dots,Z_n)-\Psi(Z^{(j)})
	\right|
	\le
	\frac{B}{n}.
	\]
	Thus changing one coordinate of the sample can change \(\Psi\) by at most \(B/n\).
	By McDiarmid's inequality, for every \(t>0\),
	\begin{align*}\textstyle
		\Pr\!\left(
		\Psi(Z_1,\dots,Z_n)
		-
		\mathbb E[\Psi(Z_1,\dots,Z_n)]
		\ge t
		\right)
		&\textstyle \le
		\exp\!\left(
		-\frac{2t^2}{\sum_{i=1}^n (B/n)^2}
		\right)  \\\textstyle
		&\textstyle =
		\exp\!\left(
		-\frac{2nt^2}{B^2}
		\right).
	\end{align*}
	Setting
	\[\textstyle
	t
	=
	B\sqrt{\frac{\log(1/\delta)}{2n}}
	\]
	yields, with probability at least \(1-\delta\),
	\begin{equation}\textstyle
		\label{eq:mcdiarmid-step}
		\Psi(Z_1,\dots,Z_n)
		\le
		\mathbb E[\Psi(Z_1,\dots,Z_n)]
		+
		B\sqrt{\frac{\log(1/\delta)}{2n}}.
	\end{equation}

	\smallskip
	\noindent
	\emph{Step 1b: Bounding the mean by Rademacher complexity.}
	It remains to bound \(\mathbb E\Psi\). Introduce an independent ghost sample
	\(Z_1',\dots,Z_n'\stackrel{\mathrm{i.i.d.}}{\sim}P_{\mathcal X}\). Since
	\[\textstyle
	\mathcal{E}(\theta)
	=
	\mathbb E_{Z'}\!\left[
	\frac1n\sum_{i=1}^n \ell(\theta;Z_i')
	\right],
	\]
	Jensen's inequality gives
	\begin{align*}\textstyle
		\mathbb E\Psi
		&\textstyle =
		\mathbb E_Z
		\left[
		\sup_{\theta\in\Theta}
		\left|
		\frac1n\sum_{i=1}^n \ell(\theta;Z_i)
		-
		\mathbb E_{Z'}\!\left[
		\frac1n\sum_{i=1}^n \ell(\theta;Z_i')
		\right]
		\right|
		\right] \\\textstyle
		&\textstyle \le
		\mathbb E_{Z,Z'}
		\left[
		\sup_{\theta\in\Theta}
		\left|
		\frac1n\sum_{i=1}^n
		\bigl(
		\ell(\theta;Z_i)-\ell(\theta;Z_i')
		\bigr)
		\right|
		\right].
	\end{align*}
	Now let \(\sigma_1,\dots,\sigma_n\) be i.i.d.\ Rademacher random variables,
	independent of \(Z\) and \(Z'\). Because \((Z_i,Z_i')\) and \((Z_i',Z_i)\) have
	the same joint law, the vector
	\[\textstyle
	\left(
	\ell(\theta;Z_i)-\ell(\theta;Z_i')
	\right)_{i=1}^n
	\]
	has the same distribution as
	\[\textstyle
	\left(
	\sigma_i
	\bigl(
	\ell(\theta;Z_i)-\ell(\theta;Z_i')
	\bigr)
	\right)_{i=1}^n.
	\]
	Hence
	\begin{align*}\textstyle
		\mathbb E\Psi
		&\textstyle \le
		\mathbb E_{Z,Z',\sigma}
		\left[
		\sup_{\theta\in\Theta}
		\left|
		\frac1n\sum_{i=1}^n
		\sigma_i
		\bigl(
		\ell(\theta;Z_i)-\ell(\theta;Z_i')
		\bigr)
		\right|
		\right] \\\textstyle
		&\textstyle \le
		\mathbb E_{Z,Z',\sigma}
		\left[
		\sup_{\theta\in\Theta}
		\left|
		\frac1n\sum_{i=1}^n
		\sigma_i\ell(\theta;Z_i)
		\right|
		\right]
		+
		\mathbb E_{Z,Z',\sigma}
		\left[
		\sup_{\theta\in\Theta}
		\left|
		\frac1n\sum_{i=1}^n
		\sigma_i\ell(\theta;Z_i')
		\right|
		\right],
	\end{align*}
	where the second inequality follows from the triangle inequality. The two terms
	are identical by symmetry, so
	\[\textstyle
	\mathbb E\Psi
	\le
	2
	\mathbb E_{Z,\sigma}
	\left[
	\sup_{\theta\in\Theta}
	\left|
	\frac1n\sum_{i=1}^n
	\sigma_i\ell(\theta;Z_i)
	\right|
	\right].
	\]
	By the definition of the Rademacher complexity used in the main text, the last
	display is exactly \(2\mathfrak R_n(\mathcal F)\). Therefore,
	\begin{equation}\textstyle
		\label{eq:expectation-step}
		\mathbb E\Psi
		\le
		2\mathfrak R_n(\mathcal F).
	\end{equation}
	Combining \eqref{eq:mcdiarmid-step} and \eqref{eq:expectation-step} proves
	\eqref{eq:generic-uniform-bound}.

	\textbf{Step 2: Applying the generic bound to the two sampling schemes.}

	\smallskip
	\noindent
	\emph{Standard i.i.d.\ sampling.}
	Under the standard protocol, the \(KT\) queried outputs
	\[\textstyle
	x_{t,k},
	\qquad
	t=1,\dots,T,\quad k=1,\dots,K,
	\]
	are i.i.d.\ samples from \(P_{\mathcal X}\). Relabel them as
	\(Z_1,\dots,Z_{KT}\). Then
	\[\textstyle
	L_{\mathrm{standard}}(\theta)
	=
	\frac1{KT}
	\sum_{k=1}^K
	\sum_{t=1}^T
	\ell(\theta;x_{t,k})
	=
	\frac1{KT}
	\sum_{i=1}^{KT}
	\ell(\theta;Z_i).
	\]
	Applying \eqref{eq:generic-uniform-bound} with \(n=KT\), we obtain that with
	probability at least \(1-\delta\),
	\[\textstyle
	\sup_{\theta\in\Theta}
	\left|
	L_{\mathrm{standard}}(\theta)-\mathcal{E}(\theta)
	\right|
	\le
	2\mathfrak R_{KT}(\mathcal F)
	+
	B\sqrt{\frac{\log(1/\delta)}{2KT}}.
	\]

	\smallskip
	\noindent
	\emph{\modelnamesimple{} sampling.}
	Under \modelnamesimple{}, the \(t\)-th access of every controlled account uses
	the same seed \(\mathrm{SG}(t)\). Since generation is deterministic once the
	latent noise is fixed, all controlled accounts receive the same output at the
	same access index:
	\[\textstyle
	x_{t,1}
	=
	x_{t,2}
	=
	\cdots
	=
	x_{t,K},
	\qquad
	t=1,\dots,T.
	\]
	Therefore,
	\begin{align*}\textstyle
		L_{\modelnamemath}(\theta)
		&\textstyle =
		\frac1{KT}
		\sum_{k=1}^K
		\sum_{t=1}^T
		\ell(\theta;x_{t,k}) \\\textstyle
		&\textstyle =
		\frac1{KT}
		\sum_{t=1}^T
		\sum_{k=1}^K
		\ell(\theta;x_{t,1}) \\\textstyle
		&\textstyle =
		\frac1T
		\sum_{t=1}^T
		\ell(\theta;x_{t,1}).
	\end{align*}
	Thus, although the distiller issues \(KT\) total queries, the empirical risk
	depends only on the \(T\) distinct outputs
	\[\textstyle
	x_{1,1},\dots,x_{T,1}.
	\]

	Because \(\mathrm{SG}\) is injective, distinct access indices are mapped to
	distinct seeds. Under the pseudorandomness assumption stated in
	Section~\ref{sec:unconditional_generation}, the resulting
	\(\epsilon_{1,1},\dots,\epsilon_{T,1}\) are independent draws from the
	target noise distribution. Since \(x=G(\epsilon)\), the corresponding outputs
	are therefore i.i.d. draws from \(P_{\mathcal X}\). Consequently,
	\[\textstyle
	x_{1,1},\dots,x_{T,1}
	\stackrel{\mathrm{i.i.d.}}{\sim}
	P_{\mathcal X}.
	\]
	Relabel \(Y_t:=x_{t,1}\), \(t=1,\dots,T\). Then
	\[\textstyle
	L_{\modelnamemath}(\theta)
	=
	\frac1T
	\sum_{t=1}^T
	\ell(\theta;Y_t),
	\qquad
	Y_1,\dots,Y_T
	\stackrel{\mathrm{i.i.d.}}{\sim}
	P_{\mathcal X}.
	\]
	Applying \eqref{eq:generic-uniform-bound} with \(n=T\), we obtain that with
	probability at least \(1-\delta\),
	\[\textstyle
	\sup_{\theta\in\Theta}
	\left|
	L_{\modelnamemath}(\theta)-\mathcal{E}(\theta)
	\right|
	\le
	2\mathfrak R_T(\mathcal F)
	+
	B\sqrt{\frac{\log(1/\delta)}{2T}}.
	\]

	Combining the two cases proves Proposition~\ref{prop:collapse}.

\section{Proof of Theorem~\ref{thm:conditional_main}}
\label{app:conditional_lads_refined}

This appendix provides the proof of Theorem~\ref{thm:conditional_main}. We
introduce the task-specific covering, bucket-level compression, center-depth
multiplicities, and population-alignment quantities needed for the analysis.
Throughout, \(N\) denotes the number of activated task-specific semantic
buckets, consistently with Section~\ref{sec:conditional_lads}.

\subsection{Task-Specific Conditional Distillation Setup}
\label{app:task_local_setup}

We use the threat model from Section~\ref{sec:setup}: over the limited
collection period, the distiller controls \(K\) accounts and each account issues
at most \(T\) queries. For account \(k\in[K]\) and within-account round
\(t\in[T]\), the distiller submits a query \(q_{t,k}\) and receives
\[\textstyle
x_{t,k}=G(q_{t,k},\epsilon_{t,k}).
\]
When convenient, we flatten the collected query-response pairs and write
\[\textstyle
\{(q_m,x_m)\}_{m=1}^{KT},
\qquad
x_m=G(q_m,\epsilon_m).
\]

We focus on task-specific distillation. Let
\[\textstyle
\mathcal Q_{\mathrm{task}} \subseteq \mathcal Q
\]
be the corresponding family of task-related prompts, and let
\(P_{\mathcal Q_{\mathrm{task}}}\) be the task-specific query distribution. The
collected queries are task-local:
\[\textstyle
q_m\in \mathcal Q_{\mathrm{task}},
\qquad
1\le m\le KT.
\]
The empirical distillation risk is
\[\textstyle
L_{\mathrm{LADS}}(\theta)
:=
\frac1{KT}\sum_{m=1}^{KT}\ell(\theta;q_m,x_m).
\]
The natural conditional population risk is
\[\textstyle
\mathcal E(\theta)
:=
\mathbb E_{q\sim P_{\mathcal Q_{\mathrm{task}}}}
\mathbb E_{x\sim P_{\mathcal X}(\cdot\mid q)}
[\ell(\theta;q,x)].
\]

The service provider fixes a private semantic hash
\[\textstyle
h:\mathcal Q\to[p]:=\{1,2,\ldots,p\}.
\]
For each account \(k\), the provider maintains a bucket-count array
\(a_k\in\mathbb N^p\).  When account \(k\) submits query \(q\), the provider
computes \(i=h(q)\), updates \(a_k[i]\leftarrow a_k[i]+1\), and sets the random
seed to
\[\textstyle
s=\mathrm{SG}(h(q),a_k[h(q)]).
\]
Thus, in the conditional setting, coupling is induced by the pair
\[\textstyle
\text{semantic bucket} + \text{bucket-local depth}.
\]
Requests from different accounts that fall into the same bucket and have the
same bucket-local depth share the same latent noise.  Requests in different
bucket-depth cells use independent latent noises.

\subsection{Task-Specific Covering and LSH-Induced Bucket Compression}
\label{app:lsh_bucket_compression}

The purpose of the task-specific covering assumption is to formalize the idea that
a multi-account distiller may submit many surface-level variants of prompts, but
the semantic complexity of a fixed task remains much smaller.

For consistency with the simplified notation in Section~\ref{sec:conditional_lads}, we use $N$ to denote the nominal
task-specific semantic complexity. In the general construction below, the number of occupied
semantic buckets may exceed $N$ due to queries that fail the center-and-hash assignment.
Proposition~\ref{prop:lsh_bucket_compression} quantifies this deviation explicitly. The simplified regime
used in Theorem~\ref{thm:conditional_main} corresponds to the clean assignment case in which these additional bad-mass
contributions vanish.

As introduced in Section~\ref{sec:conditional_lads}, the centers \(c_j\in\mathcal{Q}\) are representative queries in the query space. The theory holds for any valid task-relevant semantic metric \(d_{\mathcal{Q}}\) over \(\mathcal{Q}\), which in practical implementations may be realized through distances between dense query embeddings.

\refstepcounter{definition}\label{ass:task_local_covering}
\begin{tcolorbox}[defboxstyle, title={\textbf{Assumption \thedefinition\quad task-specific covering}}]
	There exist centers \(c_1,\ldots,c_N\in\mathcal Q\) and a radius \(R>0\) such that
	\[\textstyle
	\textstyle\Pr_{q\sim P_{\mathcal Q_{\mathrm{task}}}}
	\left(
	q\in\bigcup_{j=1}^{N}B(c_j,R)
	\right)
	\ge 1-\eta,
	\]
	where
	\[\textstyle
	B(c_j,R):=\{q\in\mathcal Q:d_{\mathcal Q}(q,c_j)\le R\},
	\]
	and \(\eta\in(0,1)\) is small.
\end{tcolorbox}

This assumption says that most task-specific queries lie near one of \(N\)
semantic centers under the task-relevant metric \(d_{\mathcal Q}\).

We next connect this covering structure to bucket-level compression
through the lens of locality-sensitive hashing
\citep{indyk1998approximate,charikar2002similarity,andoni2008near}.
Following the standard definition of
\citet{indyk1998approximate}, let $\mathcal{H}$ be an
$(R, cR, \pi_1, \pi_2)$-sensitive hash family on
$(\mathcal{Q}, d_{\mathcal{Q}})$: for any $q, q' \in \mathcal{Q}$,
\[
d_{\mathcal{Q}}(q,q') \le R \implies
\Pr_{h\sim\mathcal{H}}[h(q)=h(q')] \ge \pi_1,
\]
and
\[
d_{\mathcal{Q}}(q,q') \ge cR \implies
\Pr_{h\sim\mathcal{H}}[h(q)=h(q')] \le \pi_2,
\]
where $c>1$ and $\pi_1 > \pi_2$. Such families are known to exist for
the metrics relevant here, including Hamming and $\ell_p$ distances
\citep{indyk1998approximate,andoni2008near} and angular or cosine
distance \citep{charikar2002similarity}; the last is the natural
choice when $d_{\mathcal{Q}}$ is instantiated through distances
between normalized dense query embeddings, as in Section~\ref{sec:conditional_lads}.
The provider samples a private hash $h\sim\mathcal{H}$ at deployment
time and keeps it fixed over the collection period.

Let
\[\textstyle
B_{\mathrm{task}}:=\{h(q_m):1\le m\le KT\}
\]
be the set of occupied buckets among the collected task-specific queries.

\refstepcounter{definition}\label{prop:lsh_bucket_compression}
\begin{tcolorbox}[theoremboxstyle, title={\textbf{Proposition \thedefinition\quad LSH-induced bucket compression}}]
	Consider the \(KT\) collected query-response pairs. Suppose
	Assumption~\ref{ass:task_local_covering} holds, and suppose the private hash
	\(h\sim \mathcal H\) is sampled from an \((R,cR,\pi_1,\pi_2)\)-sensitive hash family.  Then
	there exists a nonnegative random variable \(M_{\mathrm{bad}}\) such that
	\[\textstyle
	|B_{\mathrm{task}}|\le N+M_{\mathrm{bad}},
	\]
	and
	\[\textstyle
	\mathbb E[M_{\mathrm{bad}}]\le KT(\eta+1-\pi_1).
	\]
	If the collected queries \(q_1,\ldots,q_{KT}\) are independently sampled from
	\(P_{\mathcal Q_{\mathrm{task}}}\), then for any \(\delta\in(0,1)\), with probability at least
	\(1-\delta\),
	\[\textstyle
	\textstyle|B_{\mathrm{task}}|
	\le
	N+KT(\eta+1-\pi_1)
	+\sqrt{\frac{KT}{2}\log\frac1\delta}.
	\]
\end{tcolorbox}

\begin{proof}
	For each query \(q_m\), define the bad event \(E_m\) as follows: either
	\(q_m\) does not lie in the covering union
	\(\cup_{j=1}^{N}B(c_j,R)\), or it lies in some covering ball but fails to hash
	to the same bucket as its assigned center.  More precisely, fix an assignment
	\(J(q_m)\in[N]\) whenever \(q_m\in\cup_{j=1}^{N} B(c_j,R)\) and
	\(d_{\mathcal Q}(q_m,c_{J(q_m)})\le R\).  Then \(E_m\) occurs if either the assignment does
	not exist or
	\[\textstyle
	h(q_m)\ne h(c_{J(q_m)}).
	\]
	By Assumption~\ref{ass:task_local_covering} and the near-neighbor collision
	property of LSH,
	\[\textstyle
	\Pr(E_m)\le \eta+(1-\pi_1).
	\]
	Let
	\[\textstyle
	\textstyle M_{\mathrm{bad}}:=\sum_{m=1}^{KT}\mathbf 1_{E_m}.
	\]
	Then
	\[\textstyle
	\mathbb E[M_{\mathrm{bad}}]\le KT(\eta+1-\pi_1).
	\]
	If \(q_m\) is not bad, then it hashes to one of the center buckets
	\(\{h(c_1),\ldots,h(c_N)\}\).  Therefore all occupied buckets are contained in
	the \(N\) center buckets plus the buckets contributed by bad queries, giving
	\[\textstyle
	|B_{\mathrm{task}}|\le N+M_{\mathrm{bad}}.
	\]
	When the collected queries are independent, \(M_{\mathrm{bad}}\) is a sum of
	Bernoulli variables with mean at most \(KT(\eta+1-\pi_1)\).  Hoeffding's
	inequality yields the high-probability bound.
\end{proof}

\subsection{Center-Depth Representatives}
\label{app:center_depth_representatives}

On the high-probability bucket-compression event, we call a query \(q_m\) good
if both of the following hold:
\[\textstyle
q_m\in B(c_{J(q_m)},R),
\qquad
h(q_m)=h(c_{J(q_m)}).
\]
Let \(I_{\mathrm{good}}\subseteq[KT]\) be the set of good indices and
\(I_{\mathrm{bad}}=[KT]\setminus I_{\mathrm{good}}\).  Define the empirical bad
query fraction
\[\textstyle
\textstyle\rho_\delta:=\frac{|I_{\mathrm{bad}}|}{KT}.
\]

For each good query \(q_m\), let \(j(m)\in[N]\) be its assigned center index,
so that
\[\textstyle
d_{\mathcal Q}(q_m,c_{j(m)})\le R,
\qquad
h(q_m)=h(c_{j(m)}).
\]
Let \(s(m)\in[T]\) denote the bucket-local depth of query \(q_m\) in the bucket
\(h(c_{j(m)})\).  Since each account issues at most \(T\) queries over the collection period,
\(s(m)\le T\).

For each center-depth pair \((j,s)\in[N]\times[T]\), define
\[\textstyle
\epsilon_{j,s}
\]
as the latent noise generated by seed
\[\textstyle
\mathrm{SG}(h(c_j),s),
\]
and define the corresponding center output
\[\textstyle
\widetilde x_{j,s}:=G(c_j,\epsilon_{j,s}).
\]
For every good query \(q_m\), the true output and the center output share the
same latent noise:
\[\textstyle
x_m=G(q_m,\epsilon_{j(m),s(m)}),
\qquad
\widetilde x_{j(m),s(m)}
=
G(c_{j(m)},\epsilon_{j(m),s(m)}).
\]

Define the multiplicity of each center-depth cell by
\[\textstyle
n_{j,s}
:=
\big|\{m\in I_{\mathrm{good}}:(j(m),s(m))=(j,s)\}\big|,
\]
and define the empirical weight
\[\textstyle
w_{j,s}:=\frac{n_{j,s}}{KT}.
\]
Then
\[\textstyle
\sum_{j=1}^{N}\sum_{s=1}^{T}w_{j,s}
=
\frac{|I_{\mathrm{good}}|}{KT}
=
1-\rho_\delta.
\]

The compressed weighted empirical risk is
\[\textstyle
\widetilde L(\theta)
:=
\sum_{j=1}^{N}\sum_{s=1}^{T}
w_{j,s}\,
\ell(\theta;c_j,\widetilde x_{j,s}).
\]
The corresponding weighted surrogate population target is
\[\textstyle
\widetilde{\mathcal E}(\theta)
:=
\sum_{j=1}^{N}\sum_{s=1}^{T}
w_{j,s}\,
\mathbb E_{x\sim P_{\mathcal X}(\cdot\mid c_j)}
[\ell(\theta;c_j,x)].
\]
This target is not the original conditional population risk.  It is the
surrogate population target naturally induced by the center-depth compressed
empirical process.

\subsection{Local Smoothness and Empirical Replacement Error}
\label{app:local_smoothness_replacement}

We use the following local regularity assumptions.

\refstepcounter{definition}\label{ass:task_local_smoothness}
\begin{tcolorbox}[defboxstyle, title={\textbf{Assumption \thedefinition\quad task-specific conditional smoothness}}]
	For the fixed task-specific query space \(\mathcal Q_{\mathrm{task}}\), there exists \(L_G>0\) such that for all
	\(q,q'\in \mathcal Q_{\mathrm{task}}\) and all latent noises \(\epsilon\),
	\[\textstyle
	d_{\mathcal X}(G(q,\epsilon),G(q',\epsilon))
	\le
	L_G\,d_{\mathcal Q}(q,q').
	\]
\end{tcolorbox}

\refstepcounter{definition}\label{ass:lipschitz_bounded_loss}
\begin{tcolorbox}[defboxstyle, title={\textbf{Assumption \thedefinition\quad Lipschitz and bounded distillation loss}}]
	There exist constants \(L_\ell>0\) and \(B>0\) such that for all
	\(\theta\in\Theta\) and all \((q,x),(q',x')\in \mathcal Q_{\mathrm{task}}\times \mathcal X\),
	\[\textstyle
	|\ell(\theta;q,x)-\ell(\theta;q',x')|
	\le
	L_\ell\big(d_{\mathcal Q}(q,q')+d_{\mathcal X}(x,x')\big),
	\]
	and
	\[\textstyle
	0\le \ell(\theta;q,x)\le B.
	\]
\end{tcolorbox}

\refstepcounter{definition}\label{lem:empirical_replacement}
\begin{tcolorbox}[thmboxstyle, title={\textbf{Lemma \thedefinition\quad Uniform empirical replacement by centers}}]
	On the bucket-compression event, if the empirical bad-query fraction is at most
	\(\rho_\delta\), then
	\[\textstyle
	\sup_{\theta\in\Theta}
	|L_{\mathrm{LADS}}(\theta)-\widetilde L(\theta)|
	\le
	L_\ell(1+L_G)R+B\rho_\delta.
	\]
\end{tcolorbox}

\begin{proof}
	Decompose the empirical risk into good and bad queries:
	\[\textstyle
	L_{\mathrm{LADS}}(\theta)
	=
	\frac1KT\sum_{m\in I_{\mathrm{good}}}\ell(\theta;q_m,x_m)
	+
	\frac1KT\sum_{m\in I_{\mathrm{bad}}}\ell(\theta;q_m,x_m).
	\]
	By the definition of the weights,
	\[\textstyle
	\widetilde L(\theta)
	=
	\frac1KT
	\sum_{m\in I_{\mathrm{good}}}
	\ell(\theta;c_{j(m)},\widetilde x_{j(m),s(m)}).
	\]
	Therefore, by the triangle inequality,
	\[\textstyle
	\begin{aligned}\textstyle
		\left|L_{\mathrm{LADS}}(\theta)-\widetilde L(\theta)\right|
		=
		&\textstyle \left|\frac1KT
		\sum_{m\in I_{\mathrm{good}}}
		\Big[
		\ell(\theta;q_m,x_m)
		-
		\ell(\theta;c_{j(m)},\widetilde x_{j(m),s(m)})
		\Big]+
		\frac1KT
		\sum_{m\in I_{\mathrm{bad}}}
		\ell(\theta;q_m,x_m)\right|\\\textstyle
            \leq&\textstyle  \frac{1}{KT}
            \sum_{m\in I_{\mathrm{good}}}
            \left|
            \ell(\theta;q_m,x_m)-
            \ell(\theta;c_{j(m)},\tilde{x}_{j(m),s(m)})
            \right| +
            \frac{1}{KT}
            \sum_{m\in I_{\mathrm{bad}}}
            |\ell(\theta;q_m,x_m)|.
            \end{aligned}
    	\]
	For a good query \(q_m\), the true output and representative output share the
	same latent noise:
	\[\textstyle
	x_m=G(q_m,\epsilon_{j(m),s(m)}),
	\qquad
	\widetilde x_{j(m),s(m)}
	=
	G(c_{j(m)},\epsilon_{j(m),s(m)}).
	\]
	Using Assumptions~\ref{ass:task_local_smoothness} and
	\ref{ass:lipschitz_bounded_loss},
	\[\textstyle
	\begin{aligned}\textstyle
		&\textstyle \left|
		\ell(\theta;q_m,x_m)
		-
		\ell(\theta;c_{j(m)},\widetilde x_{j(m),s(m)})
		\right|
		\\\textstyle
		&\textstyle \qquad\le
		L_\ell
		\Big(
		d_{\mathcal Q}(q_m,c_{j(m)})
		+
		d_{\mathcal X}(G(q_m,\epsilon_{j(m),s(m)}),
		G(c_{j(m)},\epsilon_{j(m),s(m)}))
		\Big)
		\\\textstyle
		&\textstyle \qquad\le
		L_\ell(1+L_G)R.
	\end{aligned}
	\]
	For bad queries, boundedness gives
	\[\textstyle
	0\le \ell(\theta;q_m,x_m)\le B.
	\]
	Averaging the good-query replacement error and the bad-query contribution gives
	\[\textstyle
	\sup_{\theta\in\Theta}
	|L_{\mathrm{LADS}}(\theta)-\widetilde L(\theta)|
	\le
	L_\ell(1+L_G)R+B\rho_\delta.
	\]
\end{proof}

\subsection{Weighted Empirical Process Bound}
\label{app:weighted_empirical_process}

The compressed empirical process is indexed by center-depth cells
\((j,s)\in[N]\times[T]\).  Its multiplicities need not be uniform.  We therefore
use a weighted empirical-process bound.

Define the weighted effective sample size associated with the unnormalized
good-query weights by
\[\textstyle
n_{\mathrm{eff}}
:=
\frac{1}{
	\sum_{j=1}^{N}\sum_{s=1}^{T}w_{j,s}^2
}.
\]
Recall that these weights satisfy
\[\textstyle
\sum_{j=1}^{N}\sum_{s=1}^{T}w_{j,s}
=
1-\rho_\delta,
\]
rather than summing to one in general.  Since at most \(NT\) weights are
nonzero, Cauchy's inequality gives
\[\textstyle
\sum_{j=1}^{N}\sum_{s=1}^{T}w_{j,s}^2
\ge
\frac{(1-\rho_\delta)^2}{NT},
\]
and therefore
\[\textstyle
n_{\mathrm{eff}}
\le
\frac{NT}{(1-\rho_\delta)^2}.
\]
In the clean-bucket case \(\rho_\delta=0\), this reduces to
\[\textstyle
n_{\mathrm{eff}}\le NT.
\]
Moreover, when the transcript is balanced over the activated center-depth
cells, so that
\[\textstyle
w_{j,s}\approx \frac{1-\rho_\delta}{NT},
\]
we have
\[\textstyle
n_{\mathrm{eff}}
\approx
\frac{NT}{(1-\rho_\delta)^2}.
\]
In the clean balanced special case,
\(\rho_\delta=0\) and hence
\[\textstyle
n_{\mathrm{eff}}\approx NT.
\]

\refstepcounter{definition}\label{lem:weighted_uniform_convergence}
\begin{tcolorbox}[thmboxstyle, title={\textbf{Lemma \thedefinition\quad Weighted uniform convergence}}]
	Let \(\mathcal F\) be a class of measurable functions taking values in
	\([0,B]\).  Let \(Z_1,\ldots,Z_m\) be mutually independent, and let
	\(w_1,\ldots,w_m\ge0\) be fixed weights.  Define
	\[\textstyle
	L_w(f):=\sum_{i=1}^{m}w_i f(Z_i),
	\qquad
	E_w(f):=\sum_{i=1}^{m}w_i\mathbb E[f(Z_i)].
	\]
	Let \(\mathbf w:=(w_1,\ldots,w_m)\). Also define the weighted Rademacher complexity
	\[\textstyle
	\mathfrak R_{\mathbf w}(\mathcal F)
	:=
	\mathbb E
	\left[
	\sup_{f\in\mathcal F}
        \left|
	\sum_{i=1}^{m}\sigma_i w_i f(Z_i)
        \right|
	\right],
	\]
	where \(\sigma_1,\ldots,\sigma_m\) are independent Rademacher random variables.
	Then, for any \(\delta\in(0,1)\), with probability at least \(1-\delta\),
	\[\textstyle
	\sup_{f\in\mathcal F}
	|L_w(f)-E_w(f)|
	\le
	2\mathfrak R_{\mathbf w}(\mathcal F)
	+
	B\sqrt{
		\frac{\log(1/\delta)}{2}
		\sum_{i=1}^{m}w_i^2
	}.
	\]
	Equivalently,
	\[\textstyle
	\sup_{f\in\mathcal F}
	|L_w(f)-E_w(f)|
	\le
	2\mathfrak R_{\mathbf w}(\mathcal F)
	+
	B\sqrt{\frac{\log(1/\delta)}{2n_{\mathrm{eff}}}}.
	\]
\end{tcolorbox}

\begin{proof}
	We give the argument for completeness, since this is the weighted analogue
		of the uniform convergence bound used in Appendix~\ref{app:proof-collapse}. Define
		\[
		\textstyle
		\Phi(Z_1,\ldots,Z_m)
		:=
		\sup_{f\in\mathcal F}
		\left|
		\sum_{i=1}^m w_i\bigl(f(Z_i)-\mathbb E f(Z_i)\bigr)
		\right|.
		\]

		First, we bound its expectation. Let \(Z_1',\ldots,Z_m'\) be an independent
		ghost sample, where \(Z_i'\) has the same distribution as \(Z_i\). By Jensen's
		inequality,
		\[
		\textstyle
		\mathbb E \Phi
		\le
		\mathbb E_{Z,Z'}
		\sup_{f\in\mathcal F}
		\left|
		\sum_{i=1}^m w_i\bigl(f(Z_i)-f(Z_i')\bigr)
		\right|.
		\]
		Introducing independent Rademacher variables
		\(\sigma_1,\ldots,\sigma_m\), and using the symmetry of
		\((Z_i,Z_i')\), we obtain
		\[
		\textstyle
		\mathbb E \Phi
		\le
		\mathbb E_{Z,Z',\sigma}
		\sup_{f\in\mathcal F}
		\left|
		\sum_{i=1}^m \sigma_i w_i\bigl(f(Z_i)-f(Z_i')\bigr)
		\right|.
		\]
		By the triangle inequality and the fact that the two ghost-sample terms have
		the same distribution,
		\[
		\textstyle
		\mathbb E \Phi
		\le
		2 \mathbb E_{Z,\sigma}
		\sup_{f\in\mathcal F}
		\left|
		\sum_{i=1}^m \sigma_i w_i f(Z_i)
		\right|.
		\]
		Equivalently, applying the same definition of weighted Rademacher complexity
		to the symmetrized class \(\mathcal F\cup(-\mathcal F)\), whose complexity is
		the same up to the displayed absolute value, gives
		\[
		\textstyle
		\mathbb E \Phi \le 2 \mathfrak R_{\mathbf w}(\mathcal F).
		\]

		It remains to concentrate \(\Phi\) around its mean. If only the \(i\)-th
		coordinate is replaced by an arbitrary \(Z_i''\), then for every
		\(f\in\mathcal F\),
		\[
		\textstyle
		\left|
		w_i(f(Z_i)-\mathbb Ef(Z_i))
		-
		w_i(f(Z_i'')-\mathbb Ef(Z_i))
		\right|
		\le B w_i,
		\]
		because \(f\in[0,B]\). Taking the supremum over \(f\) preserves this
		bounded-difference constant. Hence McDiarmid's inequality gives, for every
		\(t>0\),
		\[
		\textstyle
		\Pr\{\Phi-\mathbb E\Phi\ge t\}
		\le
		\exp\left(
		-\frac{2t^2}{B^2\sum_{i=1}^m w_i^2}
		\right).
		\]
		Setting
		\[
		\textstyle
		t
		=
		B\sqrt{\frac{\log(1/\delta)}{2}\sum_{i=1}^m w_i^2}
		\]
		yields, with probability at least \(1-\delta\),
		\[
		\textstyle
		\sup_{f\in\mathcal F}|L_w(f)-E_w(f)|
		\le
		2\mathfrak R_{\mathbf w}(\mathcal F)
		+
		B\sqrt{\frac{\log(1/\delta)}{2}\sum_{i=1}^m w_i^2}.
		\]
		Since \(n_{\mathrm{eff}}=(\sum_i w_i^2)^{-1}\), the equivalent form follows.
\end{proof}

Let
\[\textstyle
\mathcal F
:=
\{(q,x)\mapsto \ell(\theta;q,x):\theta\in\Theta\}.
\]

\refstepcounter{definition}\label{thm:weighted_compression_bound}
\begin{tcolorbox}[theoremboxstyle, title={\textbf{Theorem \thedefinition\quad task-specific weighted compression bound}}]
	Suppose Assumptions~\ref{ass:task_local_covering},
        \ref{ass:task_local_smoothness}, and
        \ref{ass:lipschitz_bounded_loss} hold. For any \(\delta\in(0,1)\), suppose further that the
        bucket-compression event described in Proposition~\ref{prop:lsh_bucket_compression}
        holds with probability at least \(1-\delta\). Conditional on the submitted queries, their bucket assignments, and
        the resulting weights $\{w_{j,s}\}$, distinct center-depth cells
        receive distinct seeds $\mathrm{SG}(h(c_j),s)$ and hence
        $\{(c_j,\tilde x_{j,s})\}$ are mutually independent with
        \[
        \tilde x_{j,s}\sim P_{\mathcal X}(\cdot\mid c_j).
        \]
         Then, with probability at
        least \(1-2\delta\),

	\[\textstyle
	\begin{aligned}\textstyle
		\sup_{\theta\in\Theta}
		|L_{\mathrm{LADS}}(\theta)-\widetilde{\mathcal E}(\theta)|
		\le\;&\textstyle
		L_\ell(1+L_G)R
		+
		B\rho_\delta
		\\\textstyle
		&\textstyle +
		2\mathfrak R_{\mathbf w}(\mathcal F)
		+
		B\sqrt{\frac{\log(1/\delta)}{2n_{\mathrm{eff}}}}.
	\end{aligned}
	\]
\end{tcolorbox}

\begin{proof}
        Let \(\mathcal A_{\mathrm{comp}}\) denote the bucket-compression event described
        in Proposition~\ref{prop:lsh_bucket_compression}. By assumption,
        \(\mathbb P(\mathcal A_{\mathrm{comp}})\ge 1-\delta\). On
        \(\mathcal A_{\mathrm{comp}}\), Lemma~\ref{lem:empirical_replacement} gives
        \[\textstyle
        \sup_{\theta\in\Theta}
        |L_{\mathrm{LADS}}(\theta)-\widetilde L(\theta)|
        \le
        L_\ell(1+L_G)R+B\rho_\delta,
        \]
        where \(\rho_\delta\) is the empirical bad-query fraction introduced in
        Section~\ref{app:center_depth_representatives}.

        Condition on the submitted queries, their bucket assignments, and hence on the
        weights \(\{w_{j,s}\}\). Under this conditioning, the weights are fixed, and
        the compressed risk is a weighted empirical process over independent
        center-depth cells:

	\[\textstyle
	\widetilde L(\theta)
	=
	\sum_{j=1}^{N}\sum_{s=1}^{T}
	w_{j,s}\ell(\theta;c_j,\widetilde x_{j,s}),
	\]
	and its weighted expectation is exactly
	\[\textstyle
	\widetilde{\mathcal E}(\theta)
	=
	\sum_{j=1}^{N}\sum_{s=1}^{T}
	w_{j,s}
	\mathbb E_{x\sim P_{\mathcal X}(\cdot\mid c_j)}
	[\ell(\theta;c_j,x)].
	\]
	Applying Lemma~\ref{lem:weighted_uniform_convergence} to the center-depth
	cells gives
	\[\textstyle
	\sup_{\theta\in\Theta}
	|\widetilde L(\theta)-\widetilde{\mathcal E}(\theta)|
	\le
	2\mathfrak R_{\mathbf w}(\mathcal F)
	+
	B\sqrt{\frac{\log(1/\delta)}{2n_{\mathrm{eff}}}}.
	\]
        The result follows by the triangle inequality and a union bound over
        \(\mathcal A_{\mathrm{comp}}\) and the weighted empirical-process
        concentration event.

\end{proof}

\subsection{Alignment with the Natural Conditional Population Risk}
\label{app:surrogate_natural_alignment}

Theorem~\ref{thm:weighted_compression_bound} controls the empirical risk around
the surrogate target \(\widetilde{\mathcal E}\).  We now align this
surrogate target with the natural conditional population risk
\[\textstyle
\mathcal E(\theta)
=
\mathbb E_{q\sim P_{\mathcal Q_{\mathrm{task}}}}
\mathbb E_{x\sim P_{\mathcal X}(\cdot\mid q)}
[\ell(\theta;q,x)].
\]
The empirical quantity \(\rho_\delta\) measures the fraction of collected
queries that fail the center-and-hash assignment used above. To compare the
surrogate target with the natural population risk, we use the corresponding
population-level bad mass under the fixed private hash.

\refstepcounter{definition}\label{ass:population_bad_mass}
\begin{tcolorbox}[defboxstyle, title={\textbf{Assumption \thedefinition\quad Population bad mass under the fixed hash}}]
	For the fixed private hash \(h\), define the good set
	\[\textstyle
	\mathcal G
	:=
	\left\{
	q\in \mathcal Q_{\mathrm{task}}:
	\exists j\in[N]\ \text{such that}\
	d_{\mathcal Q}(q,c_j)\le R
	\ \text{and}\
	h(q)=h(c_j)
	\right\}.
	\]
	Let
	\[\textstyle
	\rho
	:=
	\Pr_{q\sim P_{\mathcal Q_{\mathrm{task}}}}(q\notin\mathcal G).
	\]
	We assume \(\rho\) is small.
\end{tcolorbox}

For the concentration argument controlling empirical center masses below, we take
the collected task queries \(q_1,\ldots,q_{KT}\) to be independently sampled
from \(P_{\mathcal Q_{\mathrm{task}}}\). This condition is used only for the
alignment fluctuation in Theorem~\ref{thm:surrogate_natural_alignment}.

For every \(q\in\mathcal G\), fix an assignment
\[\textstyle
J(q)\in[N]
\]
such that
\[\textstyle
d_{\mathcal Q}(q,c_{J(q)})\le R,
\qquad
h(q)=h(c_{J(q)}).
\]
Define the population center mass
\[\textstyle
\mu_j
:=
\Pr_{q\sim P_{\mathcal Q_{\mathrm{task}}}}(q\in\mathcal G,\ J(q)=j),
\qquad
j=1,\ldots,N.
\]
Define the center-level population target
\[\textstyle
\mathcal E^{\mathrm{ctr}}(\theta)
:=
\sum_{j=1}^{N}
\mu_j
\mathbb E_{x\sim P_{\mathcal X}(\cdot\mid c_j)}
[\ell(\theta;c_j,x)].
\]
Also define the depth-aggregated empirical weight
\[\textstyle
\bar w_j
:=
\sum_{s=1}^{T}w_{j,s}.
\]
Then
\[\textstyle
\widetilde{\mathcal E}(\theta)
=
\sum_{j=1}^{N}
\bar w_j
\mathbb E_{x\sim P_{\mathcal X}(\cdot\mid c_j)}
[\ell(\theta;c_j,x)].
\]

\refstepcounter{definition}\label{thm:surrogate_natural_alignment}
\begin{tcolorbox}[theoremboxstyle, title={\textbf{Theorem \thedefinition\quad Surrogate-to-natural population alignment}}]
	Suppose Assumptions~\ref{ass:task_local_covering},
	\ref{ass:task_local_smoothness}, \ref{ass:lipschitz_bounded_loss}, and
	\ref{ass:population_bad_mass} hold, and suppose the collected task queries are
	i.i.d. from \(P_{\mathcal Q_{\mathrm{task}}}\). Then:

	\begin{enumerate}
		\item For every \(\theta\in\Theta\),
		\[\textstyle
		|\mathcal E(\theta)
		-\mathcal E^{\mathrm{ctr}}(\theta)|
		\le
		L_\ell(1+L_G)R+B\rho.
		\]

		\item For any \(\delta\in(0,1)\), with probability at least \(1-\delta\),
		\[\textstyle
		\sup_{\theta\in\Theta}
		|\widetilde{\mathcal E}(\theta)
		-\mathcal E^{\mathrm{ctr}}(\theta)|
		\le
		BN\sqrt{\frac{\log(2N/\delta)}{2KT}}.
		\]

		\item Consequently, with probability at least \(1-\delta\),
		\[\textstyle
		\sup_{\theta\in\Theta}
		|\widetilde{\mathcal E}(\theta)
		-\mathcal E(\theta)|
		\le
		L_\ell(1+L_G)R+B\rho
		+
		BN\sqrt{\frac{\log(2N/\delta)}{2KT}}.
		\]
	\end{enumerate}
\end{tcolorbox}

\begin{proof}
	First, compare the natural target with the center-level population target.
	Define
	\[\textstyle
	\psi(q,\theta)
	:=
	\mathbb E_{x\sim P_{\mathcal X}(\cdot\mid q)}
	[\ell(\theta;q,x)].
	\]
	For \(q\in\mathcal G\), let \(j=J(q)\).  Couple
	\(P_{\mathcal X}(\cdot\mid q)\) and \(P_{\mathcal X}(\cdot\mid c_j)\) through the same latent noise
	\(\epsilon\):
	\[\textstyle
	G(q,\epsilon)\sim P_{\mathcal X}(\cdot\mid q),
	\qquad
	G(c_j,\epsilon)\sim P_{\mathcal X}(\cdot\mid c_j).
	\]
	Using Assumptions~\ref{ass:task_local_smoothness} and
	\ref{ass:lipschitz_bounded_loss},
	\[\textstyle
	\begin{aligned}\textstyle
		&\textstyle |\ell(\theta;q,G(q,\epsilon))
		-
		\ell(\theta;c_j,G(c_j,\epsilon))|
		\\\textstyle
		&\textstyle \qquad\le
		L_\ell
		\big(
		d_{\mathcal Q}(q,c_j)
		+
		d_{\mathcal X}(G(q,\epsilon),G(c_j,\epsilon))
		\big)
		\le
		L_\ell(1+L_G)R.
	\end{aligned}
	\]
	Taking expectation over \(\epsilon\) gives
	\[\textstyle
	\left|
	\psi(q,\theta)
	-
	\mathbb E_{x\sim P_{\mathcal X}(\cdot\mid c_j)}
	[\ell(\theta;c_j,x)]
	\right|
	\le
	L_\ell(1+L_G)R.
	\]
	Splitting \(P_{\mathcal Q_{\mathrm{task}}}\) over \(\mathcal G\) and \(\mathcal G^c\), and using
	\(0\le\ell\le B\) on \(\mathcal G^c\), yields
	\[\textstyle
	|\mathcal E(\theta)
	-\mathcal E^{\mathrm{ctr}}(\theta)|
	\le
	L_\ell(1+L_G)R+B\rho.
	\]

	Second, define
	\[\textstyle
	\phi_j(\theta)
	:=
	\mathbb E_{x\sim P_{\mathcal X}(\cdot\mid c_j)}
	[\ell(\theta;c_j,x)].
	\]
	Then \(0\le\phi_j(\theta)\le B\), and
	\[\textstyle
	\widetilde{\mathcal E}(\theta)
	-\mathcal E^{\mathrm{ctr}}(\theta)
	=
	\sum_{j=1}^{N}(\bar w_j-\mu_j)\phi_j(\theta).
	\]
	Therefore,
	\[\textstyle
	\sup_{\theta\in\Theta}
	|\widetilde{\mathcal E}(\theta)
	-\mathcal E^{\mathrm{ctr}}(\theta)|
	\le
	B\sum_{j=1}^{N}|\bar w_j-\mu_j|.
	\]
	Under the i.i.d. query-sampling condition above, for each \(j\in[N]\),
        \[\textstyle
        \bar w_j
        =
        \frac{1}{KT}
        \sum_{m=1}^{KT}
        \mathbf 1_{\{q_m\in\mathcal G,J(q_m)=j\}}
        \]
        is an average of independent Bernoulli random variables with mean \(\mu_j\). Hence Hoeffding's inequality gives, for any \(t>0\),
        \[\textstyle
        \Pr\big(|\bar w_j-\mu_j|\ge t\big)
        \le
        2\exp(-2KTt^2).
        \]
        Taking
        \[\textstyle
        t=\sqrt{\frac{\log(2N/\delta)}{2KT}}
        \]
        and applying a union bound over \(j=1,\ldots,N\), we obtain that, with probability at least \(1-\delta\),
        \[\textstyle
        |\bar w_j-\mu_j|
        \le
        \sqrt{\frac{\log(2N/\delta)}{2KT}},
        \qquad j=1,\ldots,N.
        \]

	Hence
	\[\textstyle
	\sup_{\theta\in\Theta}
	|\widetilde{\mathcal E}(\theta)
	-\mathcal E^{\mathrm{ctr}}(\theta)|
	\le
	BN\sqrt{\frac{\log(2N/\delta)}{2KT}}.
	\]
	The third statement follows by the triangle inequality.
\end{proof}

\subsection{Conditional LADS Bound}
\label{app:conditional_lads_bound}

Combining Theorem~\ref{thm:weighted_compression_bound} and
Theorem~\ref{thm:surrogate_natural_alignment} gives a direct bound between the
empirical distillation risk and the natural conditional population risk.

\refstepcounter{definition}\label{thm:conditional_lads_complete}
\begin{tcolorbox}[theoremboxstyle, title={\textbf{Theorem \thedefinition\quad Conditional LADS bound}}]
	Under the assumptions of Theorems~\ref{thm:weighted_compression_bound} and
	\ref{thm:surrogate_natural_alignment}, for any \(\delta\in(0,1)\), with
	probability at least \(1-3\delta\),
	\[\textstyle
	\begin{aligned}\textstyle
		\sup_{\theta\in\Theta}
		|L_{\mathrm{LADS}}(\theta)-\mathcal E(\theta)|
		\le\;&\textstyle
		2L_\ell(1+L_G)R
		+
		B\rho_\delta
		+
		B\rho
		\\\textstyle
		&\textstyle +
		2\mathfrak R_{\mathbf w}(\mathcal F)
		+
		B\sqrt{\frac{\log(1/\delta)}{2n_{\mathrm{eff}}}}
		+
		BN\sqrt{\frac{\log(2N/\delta)}{2KT}}.
	\end{aligned}
	\]
\end{tcolorbox}

\begin{proof}
	For any \(\theta\in\Theta\),
	\[\textstyle
	|L_{\mathrm{LADS}}(\theta)-\mathcal E(\theta)|
	\le
	|L_{\mathrm{LADS}}(\theta)-\widetilde{\mathcal E}(\theta)|
	+
	|\widetilde{\mathcal E}(\theta)-\mathcal E(\theta)|.
	\]
	Take the supremum over \(\theta\in\Theta\), apply
	Theorem~\ref{thm:weighted_compression_bound} to the first term and
	Theorem~\ref{thm:surrogate_natural_alignment} to the second term, and combine
	the events by a union bound.
\end{proof}

\subsection{Derivation of Theorem~\ref{thm:conditional_main}}
\label{app:derivation_main_theorem}

We now connect the preceding general bound to the statement in
Theorem~\ref{thm:conditional_main}.
Proposition~\ref{prop:lsh_bucket_compression} first establishes the semantic
bucket-compression event. Conditional on this bucket structure, the remaining
probabilistic steps are the weighted empirical-process fluctuation and the
surrogate-to-natural alignment fluctuation. Accordingly, the confidence level
\(1-2\delta\) in Theorem~\ref{thm:conditional_main} accounts for these two
concentration events, while the success probability of the bucket-compression
event is quantified separately in
Proposition~\ref{prop:lsh_bucket_compression}.

The general analysis above allows for imperfect center-and-hash assignments
through the finite-sample bad-query fraction \(\rho_{\delta}\) and the
population bad mass \(\rho\). Theorem~\ref{thm:conditional_main} corresponds
to the clean-assignment specialization of this general result. In particular,
we restrict to the regime in which
\[
\rho_{\delta}=0,
\qquad
\rho=0.
\]
Hence the auxiliary terms \(B\rho_{\delta}\) and \(B\rho\) appearing in the
general bound of Theorem~\ref{thm:conditional_lads_complete} vanish.

For the center-depth empirical process, recall that the general effective
sample size is
\[
n_{\mathrm{eff}}
=
\left(
\sum_{j=1}^{N}\sum_{s=1}^{T} w_{j,s}^{2}
\right)^{-1}.
\]
Theorem~\ref{thm:conditional_main} further corresponds to the balanced
specialization of the preceding weighted bound, in which all \(NT\) active
center-depth cells carry equal empirical weight:
\[
w_{j,s}
=
\frac{1}{NT},
\qquad
j\in[N],\ s\in[T].
\]
Since
\[
\sum_{j=1}^{N}\sum_{s=1}^{T} w_{j,s}
=
1,
\]
this balanced regime is consistent with the clean-sample condition
\(\rho_{\delta}=0\). Moreover,
\[
n_{\mathrm{eff}}
=
\left(
NT\cdot \frac{1}{(NT)^2}
\right)^{-1}
=
NT,
\]
and the corresponding weighted Rademacher complexity reduces to the ordinary
\(NT\)-sample Rademacher complexity:
\[
\mathfrak{R}_{w}(\mathcal{F})
=
\mathfrak{R}_{NT}(\mathcal{F}).
\]
Therefore, the empirical-process contribution becomes
\[
2\mathfrak{R}_{NT}(\mathcal{F})
+
B\sqrt{\frac{\log(1/\delta)}{2NT}}.
\]

Under the same clean-assignment specialization, the local replacement and
population-alignment arguments contribute
\[
2L_\ell(1+L_G)R,
\]
together with the center-mass fluctuation
\[
\Delta_{\mathrm{align}}
:=
BN\sqrt{\frac{\log(2N/\delta)}{2KT}}.
\]
Combining these terms yields
\[
\begin{aligned}
\sup_{\theta\in\Theta}
\left|
L_{\mathrm{LADS}}(\theta)-\mathcal E(\theta)
\right|
\le\;&
2\mathfrak{R}_{NT}(\mathcal F)
+
B\sqrt{\frac{\log(1/\delta)}{2NT}}
\\
&+
2L_\ell(1+L_G)R
+
\Delta_{\mathrm{align}},
\end{aligned}
\]
which is exactly the bound stated in
Theorem~\ref{thm:conditional_main}.

\subsection{Reduction to the Unconditional Case}
\label{app:reduction_unconditional}

The unconditional setting is recovered as a special case.  When the query is
absent, or all queries are identical, there is only one task-specific semantic
bucket:
\[\textstyle
N=1.
\]
Moreover, there is no query-to-center replacement error, so
\[\textstyle
R=0.
\]
The compressed process then has effective sample size approximately
\[\textstyle
n_{\mathrm{eff}}\asymp T,
\]
and the dominant estimation term reduces to
\[\textstyle
2\mathfrak R_T(\mathcal F)
+
B\sqrt{\frac{\log(1/\delta)}{2T}}.
\]
Thus the conditional theory recovers the same effective sample-size collapse as
\modelnamesimple{}: although the distiller collects \(KT\) outputs, the
cross-account coupling leaves only \(T\) independent latent trajectories in the
unconditional case.  This matches the \(\mathcal O(1/\sqrt T)\) rate in
Proposition~\ref{prop:collapse}.

\end{document}